\documentclass{article} 
\usepackage{iclr2027_conference,times}

\input{math_commands.tex}

\usepackage{graphicx}
\usepackage{booktabs}
\usepackage{amssymb}
\usepackage{multirow}
\usepackage{xcolor}
\usepackage{url}
\usepackage{xurl}
\usepackage{wrapfig}
\usepackage{float}
\usepackage{enumitem}
\usepackage{hyperref}
\usepackage{svg}
\usepackage{needspace}
\usepackage[table]{xcolor}
\definecolor{g1}{HTML}{A9D18E}
\definecolor{g2}{HTML}{E2F0D9}
\newcommand{\first}[1]{\cellcolor{g1}\textbf{#1}}
\newcommand{\second}[1]{\cellcolor{g2}#1}

\floatstyle{ruled}
\newfloat{algorithm}{tbp}{loa}
\floatname{algorithm}{Algorithm}
\usepackage{algpseudocode}
\algrenewcommand\algorithmicrequire{\textbf{Input:}}
\algrenewcommand\algorithmicensure{\textbf{Output:}}
\algnewcommand{\LineComment}[1]{\State \(\triangleright\) #1}

\newcommand{\method}{AquaWAM}
\newcommand{\imagine}{\operatorname{imagine}}
\newcommand{\estimate}{\operatorname{estimate}}
\title{\method: A Dynamics-aware World Action\\ Model for Underwater Embodied Agents}

\iclrfinaltrue
\makeatletter
\def\@maketitle{\vbox{\hsize\textwidth
{\LARGE\sc \@title\par}
\vskip 0.22in
\begin{center}
{\bf Cunhao Zhu$^{1}$, Yifeng Wang$^{2}$, Dongliang Xu$^{1}$, Yunzhong Hou$^{2}$, Yue Yao$^{1}$, Chi Harold Liu$^{2}$}\\[0.45em]
{\normalsize $^{1}$Shandong University (sdu.edu.cn)\\
$^{2}$Beijing Institute of Technology (bit.edu.cn)}
\end{center}
\fancyhead{}
\renewcommand{\headrulewidth}{0pt}
\vskip 0.12in}}
\makeatother

\begin{document}
	
	\maketitle
	
	\begin{abstract}
		World Action Models (WAMs) are becoming increasingly important and useful for embodied intelligence, as they enable robots to anticipate the consequences of candidate actions before interacting with the physical environment. However, underwater robots are usually subject to passive dynamics, such as inertia, buoyancy, hydrodynamic drag, and persistent drift, which can continue to affect the vehicle even after an action is completed. Existing WAMs, which primarily predict action-conditioned visual observations, are not explicitly designed to capture such passive motion dynamics. In this paper, we present \method, the first World Action Model designed for underwater embodied agents. Instead of predicting future images, \method{} models both action-conditioned and passive physical dynamics, including the thruster dead band, the inertial glide that outlasts each command, and ambient currents. Specifically, it senses through the DVL, IMU, pressure sensor and joint encoders, while cameras supply only semantics for understanding goals and target pose. By modeling compact navigation states rather than high-dimensional visual observations, \method{} substantially reduces the model size and computational cost compared with conventional WAMs. Experimentally, \method{} achieves a 72.6\% task success rate across 20 underwater tasks on the USIM benchmark, outperforming existing methods while making action decisions 2.7$\times$ faster than U0 on an NVIDIA Jetson AGX Orin. Our model also remains effective when some onboard sensor measurements are unavailable. For example, without DVL velocity measurements, our method still achieves a 61.6\% success rate, compared with 39.4\% for U0.
		
		

	\end{abstract}
	
	\begin{figure}[h]
		\centering
		\includegraphics[width=\textwidth]{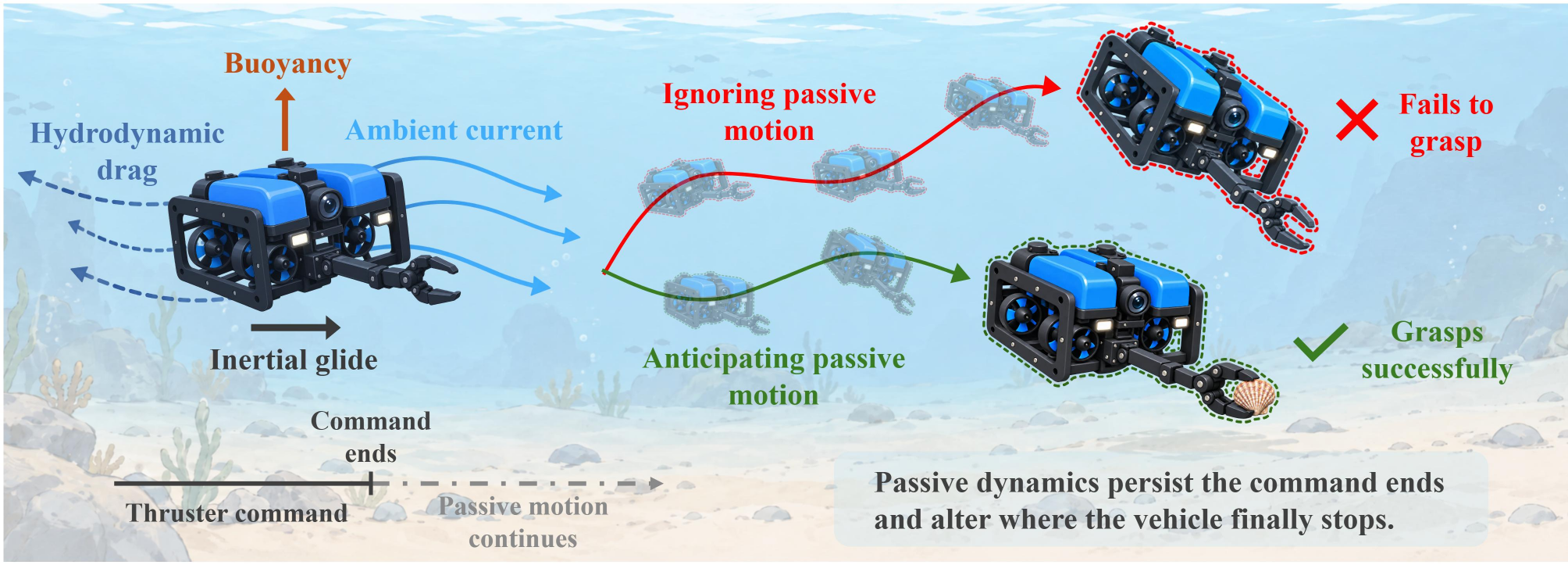}
		\caption{\textbf{%
				Motivation for a dynamics-aware World Action Model.} 
			Many existing World Action Models primarily predict future observations in video space, without explicitly modeling underwater vehicle dynamics. However, underwater motion is often strongly affected by passive physical factors such as buoyancy, drag, currents, and residual momentum, causing the vehicle state to continue evolving even after a control command is issued. Ignoring these dynamics can lead to inaccurate predictions of where and how the vehicle will move. 
		}
		\label{fig:motivation}
	\end{figure}

	\section{Introduction}
	\label{sec:intro}
	
	World Action Models (WAMs) \citep{dreamzero2026,li2025uva,cen2025worldvla} learn physical dynamics by jointly predicting the evolution of the world and the actions that drive it, enabling a robot to anticipate the consequences of a candidate action before executing it. Imitation-based policies lack such foresight, \emph{i.e.,} they reproduce the actions demonstrated by experts, whereas the consequences that determine task success are often absent from the demonstrations. Recent WAMs have demonstrated the potential of action-conditioned prediction for embodied decision making, with future observations, particularly visual observations, serving as the primary representation of how the world changes in response to an action \citep{dreamzero2026,kim2026cosmospolicy,li2026lingbotva}.
	
	
	\looseness=-1
	Underwater embodied agents, however, operate under substantially different dynamics (Figure~\ref{fig:motivation}). The state of an underwater vehicle is not determined only by its current control command. Owing to the density of water and the inertia of the hull, the vehicle continues to glide once the thrusters stop, while buoyancy and hydrodynamic drag act on it irrespective of the command. 
	Grasping, by contrast, requires the vehicle to rest within a centimeter of the object. The thrusters also exhibit a dead band, so the small corrective commands that grasping requires produce little or no thrust, and ambient currents perturb the vehicle continuously. Moreover, the Doppler velocity log (DVL) \citep{kinsey2006survey,paull2014review}, which provides velocity and, by integration, position, may lose bottom lock at any time. Since these effects are not produced by the current command, they can only be anticipated.
	
	Existing WAMs, including DreamZero \citep{dreamzero2026}, Cosmos Policy \citep{kim2026cosmospolicy} and LingBot-VA \citep{li2026lingbotva}, however, predict in video space, which is ill-suited to this setting in both computational cost and prediction target. Furthermore, video-based WAMs typically rely on large generative backbones that are costly to deploy on embedded platforms. More importantly, visual prediction provides only an indirect representation of the vehicle's future motion state, particularly under degraded underwater visibility. 
	
	
	This observation motivates a different formulation of World Action Models for underwater embodied intelligence: rather than asking \emph{what will the next image look like after this action?} an underwater agent should reason directly about \emph{how its physical state will evolve under both commanded and passive dynamics}. Such a model should capture not only the immediate response to control inputs, but also motion that persists beyond individual commands and environmental effects that occur without explicit control. At the same time, its prediction should remain sufficiently lightweight for evaluating multiple candidate actions onboard resource-constrained robotic platforms.
	
	In this paper, we present \method, the first World Action Model specifically designed for underwater embodied agents. \method{} retains the defining capability of a WAM, \emph{i.e.,} predicting the outcome of a candidate action before execution, but replaces video prediction with a compact 35-dimensional underwater physical state (Section~\ref{sec:method}) measured by the DVL, inertial measurement unit (IMU) \citep{paull2014review}, pressure sensor \citep{kinsey2006survey} and joint encoders \citep{sivcev2018manipulators}, together with the object pose relative to the gripper. 
	Camera images are excluded from the dynamics model and are used only by perception modules to estimate task goals and object poses. 
	We train \method{} on both the USIM demonstrations \citep{usim2025} and task-agnostic play data \citep{lynch2019play}; the former cover the commanded motions of the 20 tasks, while the latter expose a broader range of low-magnitude, transitional, and unstructured motions, allowing the model to learn effects such as thruster dead bands and residual glide directly from data.
	
	Experimentally, \method{} achieves a success rate of 72.6\% on the 20 USIM tasks, compared with 51.7\% for U0 \citep{usim2025}, the strongest model reported on the benchmark, with both fine-tuned on the same demonstrations under one protocol; all other  vision-language-action models (VLAs) \citep{openvla2024,pi05_2025,gr00t2025,xvla2026,smolvla2025} perform worse (Table~\ref{tab:main}). Given the object pose provided by the benchmark, the predicted thrust pulses grasp in 85.2\% of trials, compared with 17.9\% for a hand-tuned positioning law, indicating that the gain stems from prediction rather than tuning (Section~\ref{sec:results-manipulation}). When the DVL fails, the same predictor serves as the estimator that replaces it, and \method{} still succeeds in 61.6\% of trials, compared with 39.4\% for U0 (Section~\ref{sec:results-dropout}).
	
	
	Furthermore, as the model size is determined by the state dimension rather than the input resolution, \method{} is compact in both parameters and computation. It uses 2.4M parameters for the 19-dimensional vehicle state during cruising and 2.6M for the full 35-dimensional state near an object, compared with 2.7B for U0, the strongest baseline. Evaluating 128 candidate action sequences over a two-second horizon therefore takes only 22~ms on a Jetson AGX Orin~\citep{karumbunathan2022orin}, so that every thruster command is selected from predicted outcomes and re-planned every 0.5~s; a complete decision, perception included, is 2.7$\times$ faster than that of U0 on the same hardware.
	
	
	\section{Related Work}
	\label{sec:related}

	\subsection{Underwater State Estimation and Control}
	
	Underwater vehicles have long relied on model-based control, including hydrodynamic equations of motion \citep{fossen2011handbook}, thruster models that capture sluggish and dead-banded actuation \citep{yoerger1990thruster,bachmayer2000thruster}, model predictive control \citep{shen2018lmpc,heshmati2020nmpc}, and learned approaches that identify vehicle dynamics or replace the controller \citep{wehbe2017identification,carlucho2018drl}. When the DVL loses bottom lock, navigation typically falls back on inertial dead reckoning \citep{kinsey2006survey,paull2014review} or learned velocity estimation \citep{topini2020lstm,cohen2023setbeamsnet}. These approaches generally treat vehicle dynamics, task-level planning, and state estimation as separate components: the dynamics model supports control, while velocity estimation is handled by a dedicated module. In contrast, \method{} predicts the future physical state produced by each candidate action and selects commands directly from those predictions. The same predictor provides velocity when the DVL fails (Section~\ref{sec:method-estimator}), so that action prediction and navigation-state estimation share one model.
	
	\subsection{Vision-Language-Action Models}
	
	Vision-language-action (VLA) models adapt pretrained vision-language models to generate robot actions. RT-2 \citep{rt2_2023} and OpenVLA \citep{openvla2024} discretize action dimensions into tokens, whereas $\pi_0$ \citep{pi0_2024}, $\pi_{0.5}$ \citep{pi05_2025}, GR00T~N1 \citep{gr00t2025}, SmolVLA \citep{smolvla2025}, and the cross-embodiment X-VLA \citep{xvla2026} generate continuous action chunks by flow matching. USIM extends this paradigm to underwater embodied tasks, where U0 outperforms OpenVLA, $\pi_{0.5}$, and GR00T~N1.5 trained on the same demonstrations. Despite their architectural differences, these models are trained primarily to reproduce expert actions rather than to predict their consequences. Four of the six VLAs we compare against also commit to 16-step chunks, corresponding to 1.6~s of motion without intermediate outcome evaluation (Appendix~\ref{app:baselines}). This is particularly limiting underwater, where inertia, buoyancy, drag and currents keep acting on the vehicle after a command has ended. These dynamics motivate selecting actions by their predicted outcomes rather than by imitation alone.
	
	\subsection{World Models}
	
	World models provide this predictive capability by modeling how the world evolves under actions. A learned world model \citep{ha2018worldmodels} may train a policy inside imagined trajectories \citep{hafner2020dreamer,hafner2023dreamerv3} or plan at test time by rolling candidate action sequences forward \citep{chua2018pets,hafner2019planet,hansen2024tdmpc2}. Video world models predict future frames from actions \citep{hu2023gaia1,yang2023unisim,nvidia2025cosmos} or from latent actions inferred from video \citep{bruce2024genie}, while VPP \citep{hu2024vpp} and UVA \citep{li2025uva} build robot policies on video prediction models. More recent World Action Models, including DreamZero, couple future-world prediction more closely with action generation or selection, with video serving as a dense representation of the evolving world. DreamZero, for example, uses a 14B video-diffusion backbone and substantial model and system optimization to reach 7~Hz. We adopt the same predictive-action perspective, but treat video as one possible representation rather than a requirement. \method{} follows model-predictive action selection, while replacing video with the vehicle's physical state measured by its onboard sensors. This representation exposes the quantities that determine underwater control directly, including velocity, angular motion, depth, arm state, and object pose, while remaining compact enough to evaluate many candidate actions onboard.
	
	\section{\method: A Dynamics-aware World Action Model}
	\label{sec:method}
	
\begin{figure}[t]
		\centering
		\includegraphics[width=0.97\textwidth]{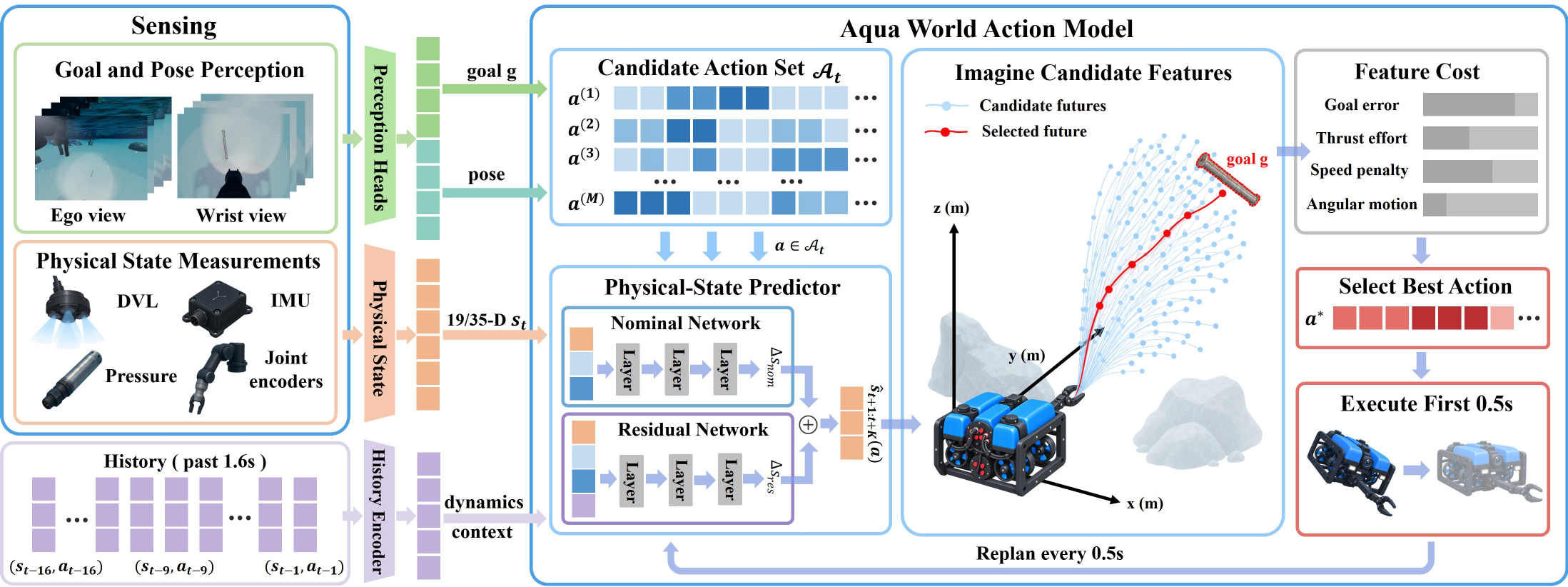}
		\caption{\textbf{Overview of \method.} Cameras supply the goal and the object pose; the predictor imagines each candidate two seconds ahead and executes the first 0.5~s of the best.}
		\label{fig:overview}
	\end{figure}

	\method{} keeps the interface of the VLAs it is compared with \citep{usim2025,pi05_2025,gr00t2025,xvla2026,smolvla2025,openvla2024}: two camera images, the DVL, IMU, pressure sensor and joint encoders, and the task instruction come in, and a chunk of thruster and joint commands goes out. What differs is how the commands are produced (Figure~\ref{fig:overview}). Given the state $s_t$ at control step $t$, the recent history $\mathcal{H}_t$ and a goal $g$,
	\begin{equation}
		\mathcal{W}_\theta:\ (s_t,\, \mathcal{H}_t,\, g)\ \longmapsto\ \big(\hat{s}^{\star}_{t+1:t+H},\ a^{\star}_{t:t+K}\big),
		\label{eq:wam}
	\end{equation}
	\method{} ($\mathcal{W}_\theta$), with parameters $\theta$, returns the future it has chosen to bring about two seconds ahead ($H = 20$ steps of 0.1~s), and the first half second of the commands $a$ that achieve it ($K = 5$ steps); the star marks the chosen candidate. A single predictor plays three roles: it predicts the physical state under any command (Section~\ref{sec:method-state}); it selects every command from these predictions, from cruising to the thrust pulses that cross the dead band (Sections~\ref{sec:method-imagination} and~\ref{sec:method-pulse}); and it replaces the DVL when it fails (Section~\ref{sec:method-estimator}). Cameras supply only goals and target pose (Section~\ref{sec:method-perception}).
	
	\subsection{Physical State and Passive Dynamics}
	\label{sec:method-state}
	
	The state consists of what the sensors read (Table~\ref{tab:state} in Appendix~\ref{app:implementation}). Body velocity and altitude from the DVL, angular rate and acceleration from the IMU, depth from the pressure sensor and the last thruster command form the compact 19-dimensional vehicle block. Within reach of an object, the joint encoders and the wrist camera add the arm and the object's pose relative to the gripper (Section~\ref{sec:method-perception}), giving 35 dimensions. Both scales share one architecture and one training recipe.
	
	A recurrent encoder $E_\phi$ with parameters $\phi$ summarizes the last 1.6~s of states and commands, $\mathcal{H}_t$, into a disturbance token $d_t = E_\phi(\mathcal{H}_t)$ that captures what no sensor measures directly, namely the ambient current and the efficiency of each thruster. A nominal network $f_{\mathrm{nom}}$ and a residual network $f_{\mathrm{res}}$ then predict the next half second of the state as an offset from the present,
	\begin{equation}
		\hat{s}_{t+1:t+K} = s_t + f_{\mathrm{nom}}(s_t, a_{t:t+K}) + f_{\mathrm{res}}(s_t, a_{t:t+K}, d_t), \qquad K = 5 \text{ steps at } 10~\mathrm{Hz}.
		\label{eq:predictor}
	\end{equation}
	Chaining four such predictions along a candidate command sequence $\mathbf{a} = a_{t:t+H}$ yields the predicted future $\hat{s}_{t+1:t+H}$, written $\imagine(s_t, \mathcal{H}_t, \mathbf{a})$. We use two seconds because it is the shortest horizon over which the hull's own response, including its overshoot, becomes visible.
	
	The predictor thus learns both the action-conditioned response and the passive dynamics that decide underwater tasks. First, the inertial glide outlasts each command: once thrust is cut, speed decays with a time constant of 2.1~s, and the offset in Eq.~\ref{eq:predictor} carries this decay even under a zero command. Second, the thrusters exhibit a dead band: commands below about 0.12 change the speed by less than 0.4~cm/s per second, whereas commands between 0.3 and 0.6 change it by about 4~cm/s per second. Third, buoyancy and drag act between commands, and the ambient current is a persistent drift that only the history reveals, hence the disturbance token. None of these effects is hand-coded; they are learned from the demonstrations and from task-agnostic play (Appendix~\ref{app:play}). Play exposes the dead band and the glide, and removing it costs the most navigation success (Section~\ref{sec:results-ablations}).
	
	\subsection{Prediction Before Action}
	\label{sec:method-imagination}
	
	The action head of a VLA usually produces the command an expert issued. Our \method{}, in contrast, has no such a action head, \emph{i.e.,} its command selects from best of 128 predicted futures, one per candidate in a set $\mathcal{A}_t$ drawn in two rounds, and the decision is revised every half second. Algorithm~\ref{alg:decision} in Appendix~\ref{app:implementation} traces one decision from the sensor readings to the executed command. The candidates are structured: a third come from a thrust-allocation law over a grid of gains, others repeat the present command or move along one axis, and the rest are snippets of play. Each candidate is rolled out two seconds ahead and scored by a evaluation function $J$, specifically, 
	\begin{equation}
		J(\mathbf{a}) = \lambda_g \sum_{k=1}^{H} \big\lVert \hat{v}_{t+k} - v_g \big\rVert + \lambda_u \sum_{k=0}^{H-1} \lVert a_{t+k} \rVert^2 + \lambda_s \Big[\max_{k} \lVert \hat{v}_{t+k} \rVert - v_{\max}\Big]_+ + \lambda_\omega \sum_{k=1}^{H} \lVert \hat{\omega}_{t+k} \rVert,
		\label{eq:score}
	\end{equation}
	where $k$ counts the predicted steps, $\hat{v}$ and $\hat{\omega}$ are the predicted body velocity and angular rate, $v_g$ is the goal velocity, $v_{\max}$ the speed cap, $[x]_+ = \max(x, 0)$, and $\lambda_g$, $\lambda_u$, $\lambda_s$ and $\lambda_\omega$ weight goal tracking, thrust, excess speed and spin. The second round is drawn around the sixteen best candidates of the first. Since the thrust law is only one candidate among many, it loses whenever the predictor foresees a glide or a dead band that the law ignores. Only the first half second of the winner is executed, and both rounds take 22~ms on a Jetson AGX Orin (Section~\ref{sec:results-baselines}). USIM's own task schema orders the sub-goals and supplies $v_g$, while every motion within it is chosen by prediction (Appendix~\ref{app:schema}).

	\subsection{Thrust Prediction for Fine Positioning}
	\label{sec:method-pulse}
	
	For grasp operation, we usually require a find positioning of robot arm. In our design, a grasp require adjust in centimeters, as the judge counts it only when the gripper closes on the object inside a window of $\pm 0.7$~cm on its tightest axis. This poses a dilemma: the corrections that reach the window are small, small commands fall inside the dead band, and any command beyond it makes the hull glide for two seconds. A continuous law must sacrifice one of the three.
	
	\method{} instead predicts discrete pulses. Within reach, the candidates become thrust pulses of direction $\mathbf{u}$, magnitude $m$ and duration $\tau$, each followed by a coast, and each is scored on where it ends rather than where it passes:
	\begin{equation}
		\begin{gathered}
			\mathcal{A}^{\mathrm{pulse}}_t = \{\text{hold}\} \cup \big\{ \text{pulse}(\mathbf{u}, m, \tau) \oplus \text{coast} \big\}, \\
			\mathbf{u} \in \big\{\boldsymbol{\delta}/\lVert\boldsymbol{\delta}\rVert,\ \operatorname{sign}(\delta_i)\,\mathbf{b}_i \ (i = x, y, z),\ -v_t/\lVert v_t \rVert\big\},\quad m \in \{0.12, \dots, 0.40\},\quad \tau \le 0.4~\mathrm{s}, \\
			J^{\mathrm{pulse}}(\mathbf{a}) = \lVert \hat{e}_{\mathrm{end}} \rVert^2_{Q} + \lambda_v \lVert \hat{v}_{\mathrm{end}} \rVert^2, \qquad Q = \mathrm{diag}\big(1/0.02^2,\; 1/0.007^2,\; 1/0.02^2\big).
		\end{gathered}
		\label{eq:pulses}
	\end{equation}
	\looseness=-1
	The magnitudes start where the recordings show the hull begins to respond, so every thrust pulse crosses the dead band by construction. Here hold lets the hull glide without translational thrust, $\oplus$ denotes concatenation in time, $\boldsymbol{\delta}$ is the body-frame displacement from the gripper to its target point above the object, $\mathbf{b}_i$ are the body axes, $-v_t/\lVert v_t \rVert$ brakes against the current velocity, $\hat{e}_{\mathrm{end}}$ is the predicted gripper--object offset $e = (x, y, z)$ over the last third of the horizon, $\hat{v}_{\mathrm{end}}$ the predicted velocity at its last step, $\lVert e \rVert^2_{Q} = e^{\top} Q e$ with $Q$ making the $\pm 0.7$~cm tolerance on $y$ dominate, and $\lambda_v$ weights coming to rest (Appendix~\ref{app:scores}). The pulse whose predicted glide ends inside the window at rest is selected, and because the last step of every candidate is a coast, the executed motion matches the predicted one even if the next decision is late. The hull acts as the fine manipulator: the arm's lateral reach is only $\pm 3$~cm, so the arm holds a ready pose and the hull does the centimeter-level positioning (Appendix~\ref{app:schema}). The gripper, the only discrete command, also closes by prediction, when an outcome model trained on 3{,}488 recorded closures predicts a firm hold with probability at least 0.7.
	
	\subsection{Semantic Decoding for Objectives}
	\label{sec:method-perception}
	
	Camera images are not directly used by the predictor. Instead, we have a perception heads, \emph{i.e.,} a DINOv2-base backbones \citep{oquab2024dinov2} fine-tuned on USIM, convert both camera images and the instruction into the quantities the predictor works with, meters and meters per second. A navigation head predicts the displacement to the current waypoint three seconds ahead, which serves as the goal, and a wrist-camera head predicts the target pose, \emph{i.e.,} the object's position and orientation in the gripper frame with a per-axis uncertainty, and locates the destination box for transport. Each of USIM's nine instructions names a task and a goal type. The heads are trained on the demonstrations and on \method{}'s own rollouts relabeled with the reference goals \citep{ross2011dagger}, so that they see the states \method{} visits rather than only those of the experts, while the predictor stays fixed.
	
	\subsection{\method{} as a Navigation-State Estimator}
	\label{sec:method-estimator}
	
\begin{wrapfigure}{r}{0.52\textwidth}
		\vspace{-0.6\baselineskip}
		\centering
		\includegraphics[width=\linewidth]{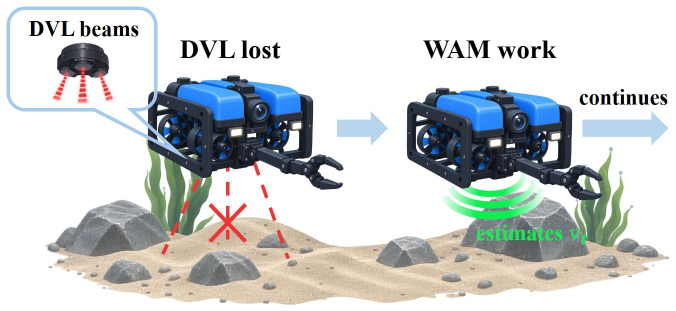}
		\caption{\textbf{\method{} as a navigation-state estimator.} After DVL loss, it can still estimate the velocity.}
		\label{fig:estimator}
	\end{wrapfigure}

	We further simulate a case when the DVL reports a loss of bottom lock, the state loses velocity and, through it, position. Vehicles usually carry a separate module for this, such as an inertial filter or a learned estimator. \method{} does not: prediction and estimation are two readings of the same network (Figure~\ref{fig:estimator}),
	\begin{equation}
		\begin{gathered}
			\imagine:\ (s_t,\ \mathcal{H}_t,\ \mathbf{a}) \mapsto \hat{s}_{t+1:t+H}, \\
			\estimate:\ \mathcal{H}_t^{v} \mapsto \hat{v}_t .
		\end{gathered}
		\label{eq:dual}
	\end{equation}
	Estimation, the second reading in Eq.~\ref{eq:dual}, masks the velocity channels of the history, $\mathcal{H}_t^{v}$, recomputes the disturbance token from the remaining channels, and asks three velocity heads trained on such masked histories what the IMU readings and the commands imply; their mean is the estimate and their spread is $\sigma_t$. The estimate takes the place of the DVL, the IMU attitude and the pressure sensor complete the dead-reckoned pose, and the target pose is re-anchored in that frame. Since a blind estimate should not be trusted equally at all times, a cumulative statistic $S_t$ accumulates the deviation $z_t$ of the estimate from its value $v_0$ at the moment of loss, in units of the spread $\sigma_t$ plus a noise floor $\sigma_0$, with a forgetting factor $\gamma$ and a slack $\kappa$,
	\begin{equation}
		z_t = \frac{\lVert \hat{v}_t - v_0 \rVert}{\lVert \sigma_t + \sigma_0 \rVert}, \qquad
		S_t = \max\big(0,\ \gamma S_{t-1} + z_t - \kappa\big), \qquad
		\alpha_t = \min(1,\ S_t / \eta).
		\label{eq:trust}
	\end{equation}
	
	The trust $\alpha_t$ reaches one when $S_t$ reaches the level $\eta$, and it bounds how far the new decision may depart from the last one made with the DVL. Both $\kappa$ and $\eta$ are calibrated on held-out blind windows to a chosen false-alarm rate, and a known change of goal makes trust immediate. When cruising toward a fixed goal, \method{} first brakes into the speed range where the estimate is most accurate, while an attitude guard keeps it level. The estimator is retrained on the vehicle's own blind flights and agrees with the predictor by construction, since it reports the velocity the predictor would forecast for the executed command. \method{} therefore steers blind with the same physics it predicts with (Section~\ref{sec:results-dropout}).
	
	\section{Experiments}
	\label{sec:results}
	
\begin{figure}[t]
		\centering
		\includegraphics[width=\textwidth]{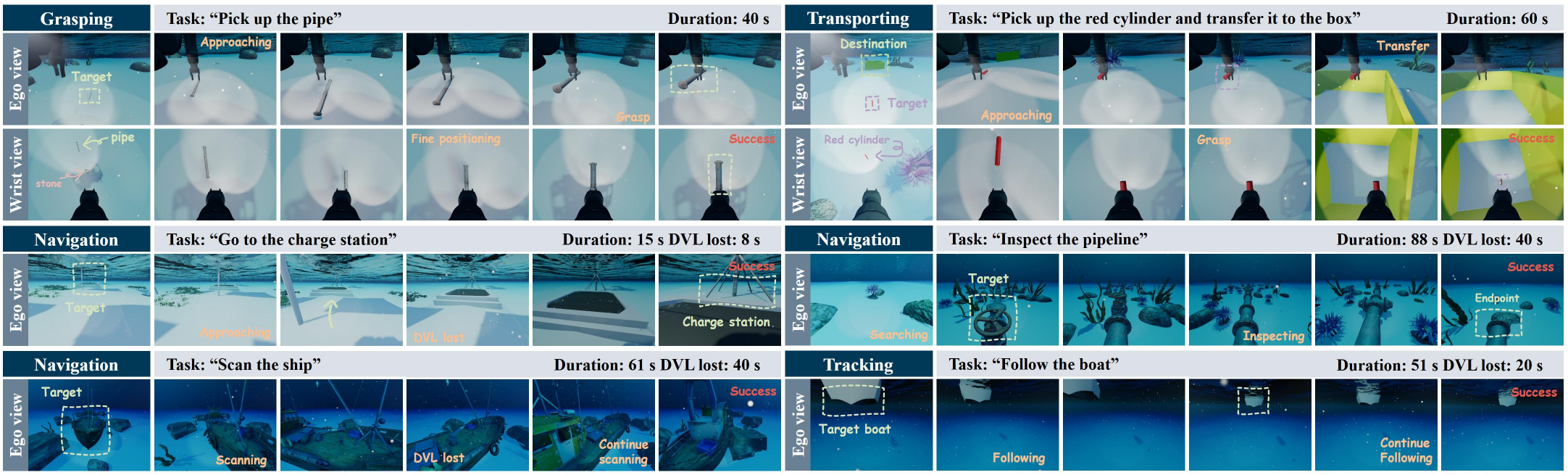}
		\vspace{-2em}
		\caption{\textbf{Execution traces of \method.} Grasping and transporting run with full sensing; navigation and tracking lose the DVL where marked and finish on the estimate.}
		\label{fig:traces}
		\vspace{-1.2em}
	\end{figure}

	We evaluate on USIM, the benchmark introduced together with U0. It simulates a BlueROV2-class vehicle with an arm in Stonefish \citep{cieslak2019stonefish} and comprises 20 tasks in four families (6 navigation, 12 grasping, 1 transporting, 1 tracking), with 700 trials per model allocated as the benchmark prescribes (160, 480, 40 and 20). Every model sees the same two cameras, issues the same 13 commands and is scored by the same judges. Under \emph{DVL dropout}, the DVL loses bottom lock partway through every trial and never recovers (8~s into a goto, 20~s into tracking, 30~s into a grasp or transport, 40~s into an inspection or scan), with unchanged judges. Figure~\ref{fig:traces} shows examples.
	
	All baselines, namely U0, $\pi_{0.5}$, GR00T~N1.5, X-VLA, SmolVLA and OpenVLA, are fine-tuned on the USIM demonstrations (Appendix~\ref{app:baselines}). We report successes per family and the overall success rate (SR), together with the quality metrics as defined by USIM: success weighted by path length (SPL) \citep{anderson2018spl}, average success duration (ASD) and stage success rate (SSR).
	
	\subsection{Main Results}
	\label{sec:results-main}
	
	\suppressfloats[t]
	\setlength{\floatsep}{8pt plus 2pt minus 2pt}
	\setlength{\textfloatsep}{12pt plus 2pt minus 3pt}
	
	\begin{table}[t]
		\caption{\textbf{Results on the 20 USIM tasks.} Succ. denotes the number of successes trials; SPL and ASD follow the USIM definitions. The {\setlength{\fboxsep}{1pt}\colorbox{g1}{\textbf{best}}} and {\setlength{\fboxsep}{1pt}\colorbox{g2}{second-best}} are highlighted.}
		\label{tab:main}
		\centering
		\scriptsize
		\setlength{\tabcolsep}{4pt}
		\renewcommand{\arraystretch}{1.08}
		\resizebox{\textwidth}{!}{%
			\begin{tabular}{c cc cc cc c c}
				\toprule
				\multirow{2}{*}[-2.5pt]{Model} & \multicolumn{2}{c}{Navigation} & \multicolumn{2}{c}{Grasping} & \multicolumn{2}{c}{Transporting} & Tracking & Overall \\
				\cmidrule(lr){2-3}\cmidrule(lr){4-5}\cmidrule(lr){6-7}\cmidrule(lr){8-8}\cmidrule(lr){9-9}
				& Succ.$\uparrow$ & SPL$\uparrow$ & Succ.$\uparrow$ & ASD (s)$\downarrow$ & Succ.$\uparrow$ & SSR$\uparrow$ & Succ.$\uparrow$ & SR$\uparrow$ (succ./trials) \\
				\midrule
				OpenVLA & 50/160 & 0.71$\pm$0.19 & 0/480 & -- & 0/40 & 0.0\% & 0/20 & 7.1\% (50/700) \\
				$\pi_{0.5}$ & 58/160 & 0.59$\pm$0.19 & 1/480 & 87.3 & 0/40 & 0.0\% & 0/20 & 8.4\% (59/700) \\
				GR00T N1.5 & 122/160 & \second{0.78$\pm$0.23} & 140/480 & 102.2 & 10/40 & 32.5\% & 4/20 & 39.4\% (276/700) \\
				X-VLA & 85/160 & 0.64$\pm$0.24 & 4/480 & \second{63.5} & 0/40 & 0.0\% & 1/20 & 12.9\% (90/700) \\
				SmolVLA & 74/160 & 0.64$\pm$0.25 & 7/480 & 94.7 & 0/40 & 5.0\% & 1/20 & 11.7\% (82/700) \\
				U0 & \second{140/160} & 0.75$\pm$0.24 & \second{195/480} & 99.8 & \second{13/40} & \second{52.5\%} & \second{14/20} & \second{51.7\% (362/700)} \\
				\midrule
				\method{} (Ours) & \first{151/160} & \first{0.84$\pm$0.21} & \first{314/480} & \first{62.9} & \first{23/40} & \first{65.0\%} & \first{20/20} & \first{72.6\% (508/700)} \\
				\bottomrule
			\end{tabular}%
		}
		\vspace{-1em}
	\end{table}
	
	Under full sensing, \method{} succeeds in 72.6\% of the 700 trials, compared with 51.7\% for U0 (Table~\ref{tab:main}). The margin holds in every family and is largest where the water matters most: grasping improves from 40.6\% to 65.4\%, transporting from 32.5\% to 57.5\% and tracking from 70\% to 100\%, while navigation rises from 87.5\% to 94.4\%. GR00T~N1.5, U0's backbone, reaches 39.4\%.

	The quality metrics follow the success rates: \method{}'s successful grasps take 62.9~s compared with 99.8~s for U0, and its navigation paths are the most efficient (SPL 0.84).
	
	\subsection{Grasp Experiments}
	\label{sec:results-manipulation}
	
	\begin{wrapfigure}{r}{0.44\textwidth}
		\vspace{-0.4\baselineskip}
		\centering
		\includegraphics[width=\linewidth]{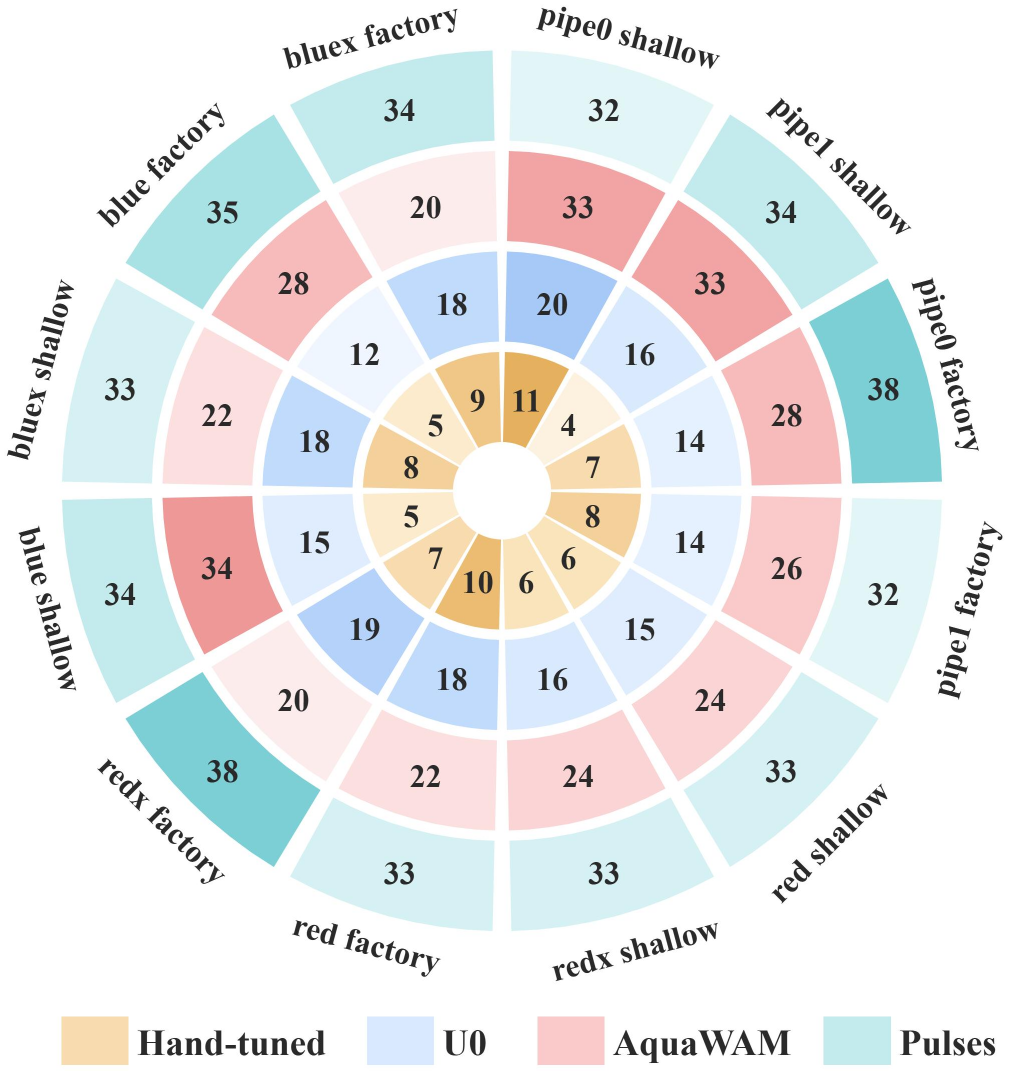}
		\vspace{-1.5\baselineskip}
		\caption{\textbf{Experiments on different grasp tasks.} We show the number of successes over 40 per task.}
		\label{fig:grasp}
		\vspace{-0.8\baselineskip}
	\end{wrapfigure}
	Figure~\ref{fig:grasp} breaks the grasping margin down by task: U0 succeeds in 12 to 20 of 40 trials per object and \method{} in 20 to 34, outperforming U0 on every object. To identify its source, we fix everything except the decision rule, using the same task program, closure thresholds and the benchmark's own object pose, so that perception plays no part. In this setting, the hand-tuned positioning law of Section~\ref{sec:method-pulse} grasps in 17.9\% of the 480 trials and the predicted pulses in 85.2\%, the inner and outer rings of Figure~\ref{fig:grasp}. The law already fails during the approach: only 24\% of its trials bring the gripper inside the closure window, while the rest dither in the dead band.
	
	We further rule out three alternative explanations on the same tasks. For cadence, at \method{}'s half-second rhythm the law reaches 12/40 and 11/40 on two tasks (11 and 6 at 1.6~s), against 32/40 and 33/40 for the pulses. For the trigger, the learned closure model turns 32/40 and 33/40 into 35/40 and 34/40, so the moment of closing does not explain the margin. For the grid, pulse grids of $2\times2$, $5\times4$ (deployed) and $9\times4$ give 34, 32 and 31 of 40. The margin thus requires a predictor that knows where a pulse ends. \method{} itself reads the object pose from the wrist camera and grasps in 65.4\% (Table~\ref{tab:main}), so the remaining gap to 85.2\% lies in perception.
	
	\subsection{When the DVL Fails}
	\label{sec:results-dropout}

	\begin{figure}[t]
		\centering
		\includegraphics[width=\linewidth]{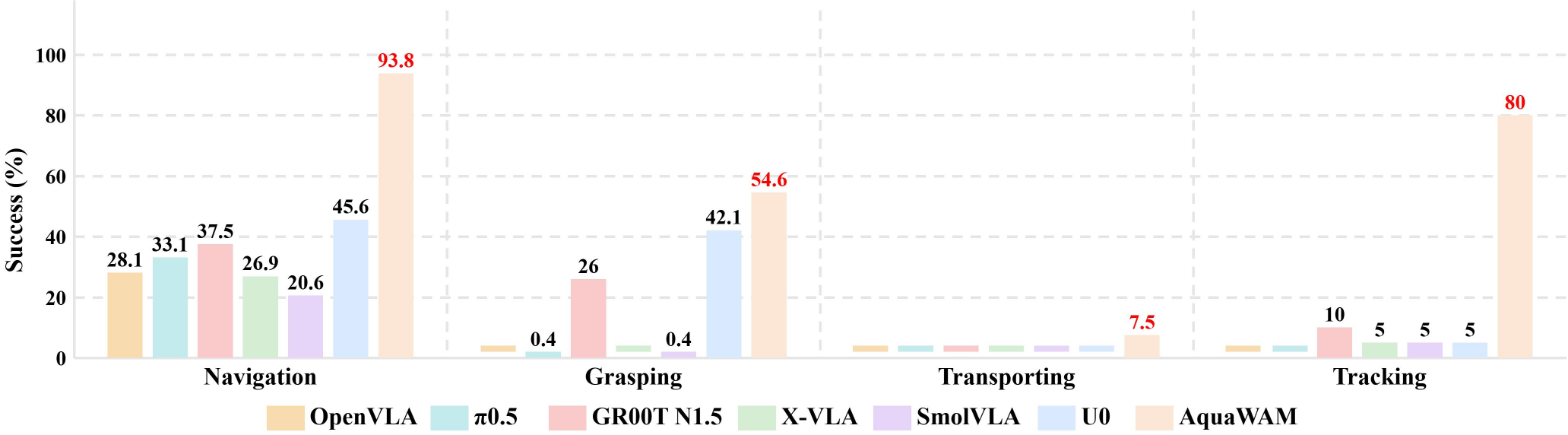}
		\vspace{-2em}
		\caption{\textbf{Success under DVL dropout, per task family.} The DVL loses bottom lock partway through every trial and never recovers. Colored baseline marks denote zero success.}
		\label{fig:dropout}
	\end{figure}
	
	When the DVL loses bottom lock, the models of Table~\ref{tab:main} separate (Figure~\ref{fig:dropout}). U0 falls from 51.7\% to 39.4\%: its navigation halves (87.5\% to 45.6\%), its tracking nearly disappears (70\% to 5\%), and only grasping, which relies on sight, holds at 42.1\%. GR00T~N1.5 falls from 39.4\% to 26.7\%. \method{} retains 61.6\%: navigation 93.8\% and tracking 80\% on dead reckoning, and grasping 54.6\%, where the approach runs on the estimate until the wrist camera re-anchors the object. Transporting, which carries an object 2.9~m after the loss, remains low for both (7.5\% against 0\%).
	
	The estimator also transfers to other policies. Used as a takeover layer that stands in for U0's DVL once it reports loss, it restores U0's navigation (96.9\%) and most of its tracking (60\%), raising U0 from 39.4\% to 49.7\%; the remaining twelve points to \method{} lie in grasping and tracking.
	
	\subsection {Experiments on Extreme Cases}
	\label{sec:results-hard}
	
	\begin{table}[t]
		\vspace{-1em}
		\caption{\textbf{Results on extreme cases, \emph{i.e.,} USIM-Hard.} Left: perturbed worlds, 20 trials per task. Right: stricter judges re-score the trials of Table~\ref{tab:main}, and in the DVL-loss rows those of Section~\ref{sec:results-dropout}.}
		\label{tab:hard}
		\centering
		\scriptsize
		\setlength{\tabcolsep}{2pt}
		\renewcommand{\arraystretch}{1.08}
		\begin{tabular*}{\textwidth}{@{\extracolsep{\fill}}l cc @{\hspace{6pt}} l cc@{}}
			\toprule
			Perturbed world & U0 & \method & Stricter judge & U0 & \method \\
			\midrule
			Start heading rotated 90--180$^\circ$ 
			& 52/80 & 63/80 
			& Goto, all waypoints within 1.0 m 
			& 53/80 & 61/80 \\
			
			Goto endpoint within 0.5 m, not 1 m 
			& 26/40 & 35/40 
			& Goto, all waypoints within 0.5 m 
			& 44/80 & 44/80 \\
			
			Thruster 3 at 50\,\%, seen in play 
			& 3/40 & 30/40 
			& Goto, 0.5 m waypoints, DVL loss 
			& 28/80 & 45/80 \\
			
			Thruster 1 at 50\,\%, unseen 
			& 4/40 & 21/40 
			& Goto, all waypoints within 0.25 m 
			& 25/80 & 24/80 \\
			
			Thruster 1 at 0\,\%, unseen 
			& 1/40 & 4/40 
			& Inspect, 1 m ball, 90\,\% coverage 
			& 36/40 & 39/40 \\
			
			Turbid water, Jerlov 0.50 
			& 1/20 & 9/20 
			& Inspect, same judge, DVL loss 
			& 7/40 & 39/40 \\
			\bottomrule
		\end{tabular*}
		
	\end{table}
	
	Since the official judges saturate near the top of Table~\ref{tab:main}, we tighten them and perturb the world (Table~\ref{tab:hard}). Under stricter judges, \method{} passes every waypoint within 1.0~m in 61 of 80 gotos against 53 for U0, the two are level at 0.5~m with 44 each, and U0 keeps 25 against 24 at 0.25~m. This is expected, since \method{} advances once it is within 0.9~m of a waypoint (Appendix~\ref{app:schema}) and is never asked to pass closer; nevertheless, on the harder water-tower route it leads at every tolerance (Appendix~\ref{app:hard}). With the DVL lost, \method{} still passes the 0.5~m waypoints in 45 of 80 gotos against 28 and completes 39 of 40 inspections against 7. Under perturbation, \method{} outperforms U0 on every protocol, by 27 of 40 with thruster 3 at half thrust and by 8 of 20 in turbid water, and with thruster 1 delivering no thrust the two gotos drop to 4 and 1 of 40 (Appendix~\ref{app:hard}).
	
	\subsection{Ablations}
	\label{sec:results-ablations}
	
	\begin{table}[t]
		\caption{\textbf{Ablation studies.} Successes are given as succeeded/trials; the last two rows run when DVL is not available. SPL (mean $\pm$ s.d.) and ASD (s) follow definitions from USIM dataset. We simulate a case when DVL fails and alternatively use AquaWAM/IMU to predict/calculate velocity.  }
		\label{tab:ablation}
		\centering
		\scriptsize
		\setlength{\tabcolsep}{4pt}
		\renewcommand{\arraystretch}{1.08}
		\resizebox{\textwidth}{!}{%
			\begin{tabular}{c cc cc cc c c}
				\toprule
				\multirow{2}{*}[-2.5pt]{Model} & \multicolumn{2}{c}{Navigation} & \multicolumn{2}{c}{Grasping} & \multicolumn{2}{c}{Transporting} & Tracking & Overall \\
				\cmidrule(lr){2-3}\cmidrule(lr){4-5}\cmidrule(lr){6-7}\cmidrule(lr){8-8}\cmidrule(lr){9-9}
				& Succ.$\uparrow$ & SPL$\uparrow$ & Succ.$\uparrow$ & ASD (s)$\downarrow$ & Succ.$\uparrow$ & SSR$\uparrow$ & Succ.$\uparrow$ & SR$\uparrow$ (succ./trials) \\
				\midrule
				\textbf{\method{}} & \textbf{151/160} & \textbf{0.84$\pm$0.21} & \textbf{314/480} & \textbf{62.9} & \textbf{23/40} & \textbf{65.0\%} & \textbf{20/20} & \textbf{72.6\% (508/700)} \\
				w/o imagination & 127/160 & 0.69$\pm$0.22 & 79/480 & 93.7 & 3/40 & 17.8\% & 14/20 & 31.9\% (223/700) \\
				w/o play data & 81/160 & 0.53$\pm$0.24 & 138/480 & 97.3 & 7/40 & 26.3\% & 13/20 & 34.1\% (239/700) \\
				w/o disturbance token & 89/160 & 0.59$\pm$0.23 & 193/480 & 81.7 & 11/40 & 41.7\% & 12/20 & 43.6\% (305/700) \\
				\midrule
				\textbf{\method{} as DVL} & \textbf{150/160} & \textbf{0.82$\pm$0.21} & \textbf{262/480} & \textbf{65.8} & \textbf{3/40} & \textbf{67.5\%} & \textbf{16/20} & \textbf{61.6\% (431/700)} \\
				IMU as DVL & 86/160 & 0.49$\pm$0.26 & 93/480 & 103.6 & 1/40 & 13.2\% & 4/20 & 26.3\% (184/700) \\
				\bottomrule
			\end{tabular}%
		}
		\vspace{-5pt}
	\end{table}
	
	Table~\ref{tab:ablation} removes the key components of \method{} one at a time on the same 700 trials. Without imagination, \emph{i.e.,} with a single forward pass of an amortized head in place of the predicted rollouts, grasping falls from 65.4\% to 16.5\% and transporting from 57.5\% to 7.5\%, while navigation remains at 79.4\%, indicating that the command crossing the dead band must be predicted. Without the play data, the predictor never observes the glide or the dead band, and navigation drops to 50.6\% with SPL 0.53. Without the disturbance token, the predictor cannot distinguish the ambient current from its own motion, and overall success falls to 305 of 700. With the IMU alone standing in for the DVL, success falls to 184 of 700 trials, against 431 with \method{} as DVL, and tracking falls from 16/20 to 4/20. Each component helps most where the dynamics it models matter.
	
	\subsection{Computation Costs}
	\label{sec:results-baselines}
	
	\begin{wrapfigure}{r}{0.46\textwidth}
		\vspace{-1.5\baselineskip}
		\centering
		\includegraphics[width=\linewidth]{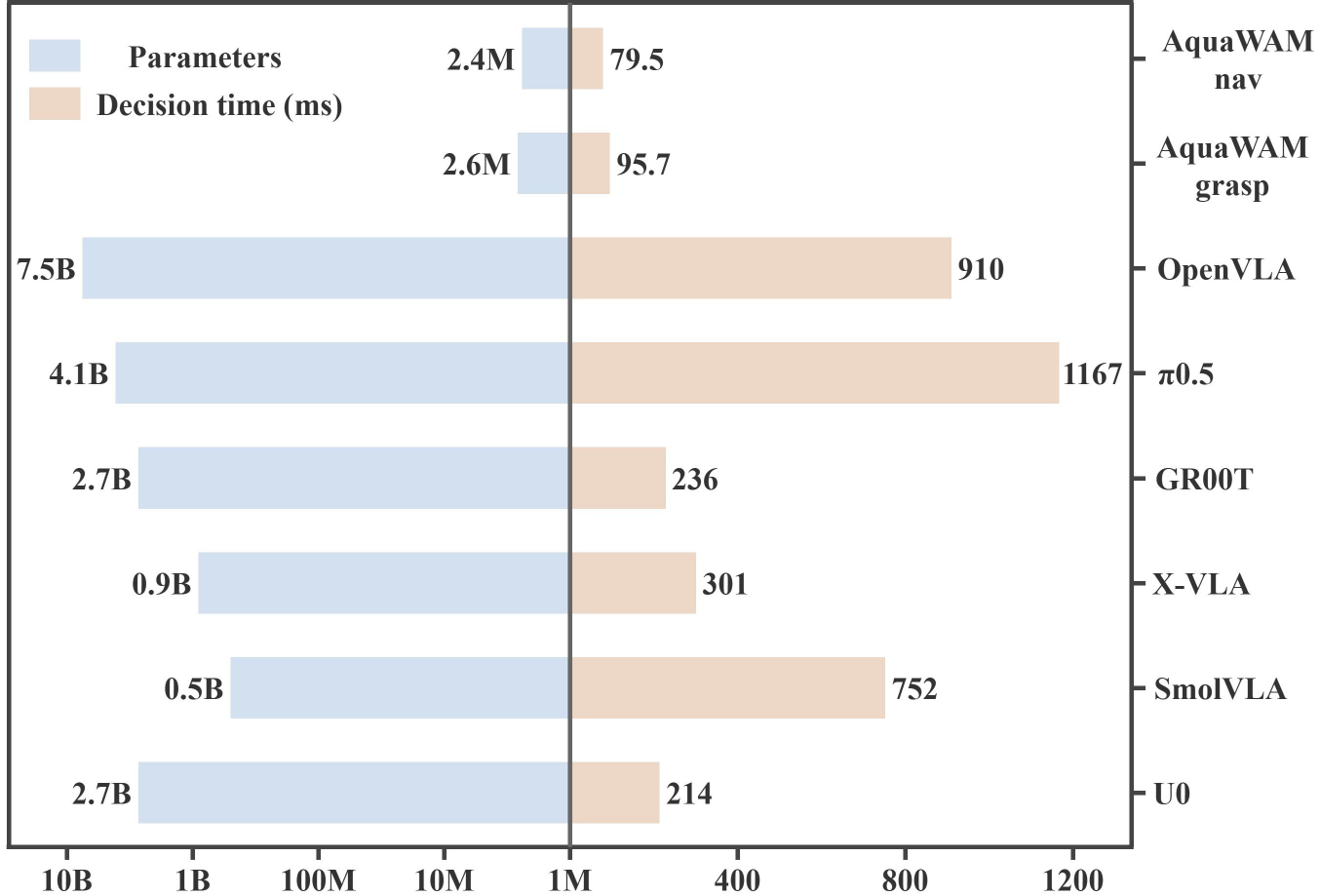}
		\vspace{-1.95\baselineskip}
		\caption{\textbf{Computation cost on Jetson AGX Orin 32~GB.} We show the number of parameters (left) and the running time (right).}
		\label{fig:cost}
		\vspace{-1\baselineskip}
	\end{wrapfigure}
	\method{} is designed to run on the vehicle. On a Jetson AGX Orin, evaluating 128 candidates two seconds ahead in two rounds takes 22~ms and a pulse decision 19~ms. Including the perception heads, a decision costs 79.5~ms on a navigation step and 95.7~ms on a grasp step, measured end to end on replayed sensor streams, which is 2.7$\times$ faster than U0's 214~ms for each 1.6~s action chunk. Per inference call, GR00T~N1.5 needs 236~ms, X-VLA 301, SmolVLA 752, OpenVLA 910 and $\pi_{0.5}$ 1167 (Figure~\ref{fig:cost}). The 2.4M and 2.6M parameters belong to the predictor; the DINOv2-base backbones of the perception heads add 86M each, and their time is included in every \method{} figure above.
	
	\section{Conclusion}
	\label{sec:conclusion}
	
	In this paper, we presented \method, the first World Action Model for underwater embodied agents. \method{} retains the defining capability of a WAM, predicting what a candidate action will do before acting, but changes what is predicted: a 35-dimensional physical state read from the vehicle's own sensors, whose passive dynamics are learned from task-agnostic play. This design yields three benefits. First, the model runs on the vehicle's own computer, taking 79.5~ms per decision on a Jetson AGX Orin. Second, its predicted thrust pulses cross the dead band and grasp within a centimeter where a hand-tuned law fails. Third, the same predictor replaces a failed DVL, both for itself and as a takeover layer for a VLA. 
	
	\paragraph{Limitations.} All results are obtained in simulation: USIM models the hydrodynamics that \method{} learns from, but a real deployment adds effects it only approximates, such as sensor noise, latency and intermittent DVL dropouts. Although \method{} outperforms the other models, it still struggles in three cases. Transport after DVL loss remains at 7.5\%, since the object must still be carried 2.9~m on an estimated velocity; a thruster that delivers no thrust, a fault absent from the play, leaves only 4 of 40 gotos successful; and perception remains a bottleneck: the same pulses grasp in 85.2\% of trials with the benchmark's object pose but 65.4\% from the wrist camera.
	
	\paragraph{Future work.} Our next step is to deploy \method{} on a physical remotely operated vehicle (ROV) \citep{sivcev2018manipulators}: the predictor reads only sensors that every ROV carries, so it can be adapted with unlabeled play data from the target hull, and the disturbance token, which already compensates for a weakened thruster (Table~\ref{tab:hard}), offers a mechanism for online adaptation. We further plan to feed visual cues into the navigation-state estimator so that the cameras that supply the goals also bound the dead-reckoning drift that currently limits transport after DVL loss.

	
	\subsection*{AI Use Statement}
	
	Generative AI tools were used only to polish the writing of this paper, that is, grammar, wording and readability. They were not used to generate data, formulate claims, design or run experiments, implement methods, or analyze and interpret results. The authors reviewed every edit and take full responsibility for the content of the paper.
	
	\subsection*{Ethics Statement}
	
	This work studies underwater robot control in simulation only, on the USIM benchmark and the open-source Stonefish simulator. It involves no physical vehicle, no human subjects and no personal data. Deploying a learned controller on a real vehicle would require safety validation beyond the scope of this paper.
	
	\subsection*{Reproducibility Statement}
	
	The predictor, the score weights, the hand-tuned positioning law, the task list, play-data collection, baseline fine-tuning, the per-task results, and the failure cases are documented in Appendices~\ref{app:implementation}--\ref{app:failures}. Supplementary videos show representative episodes of every model (Appendix~\ref{app:qualitative}). The code, the play data and the trained models will be released with the paper.
	
	\bibliographystyle{iclr2027_conference}
	\bibliography{waterwam}

@article{usim2025,
	title   = {{USIM} and {U0}: A Vision-Language-Action Dataset and Model for General Underwater Robots},
	author  = {Gu, Junwen and Wu, Zhiheng and Si, Pengxuan and Qiu, Shuang and Zhang, Zhentao and Feng, Yukai and Sun, Luoyang and Luo, Laien and Yu, Lianyi and Wang, Jian and Wu, Zhengxing},
	journal = {arXiv preprint arXiv:2510.07869},
	year    = {2025},
	note    = {Version 4, May 2026}
}

@inproceedings{rt2_2023,
	title     = {{RT-2}: Vision-Language-Action Models Transfer Web Knowledge to Robotic Control},
	author    = {Brohan, Anthony and others},
	booktitle = {Conference on Robot Learning (CoRL)},
	series    = {Proceedings of Machine Learning Research},
	volume    = {229},
	pages     = {2165--2183},
	publisher = {PMLR},
	year      = {2023}
}

@inproceedings{openvla2024,
	title     = {{OpenVLA}: An Open-Source Vision-Language-Action Model},
	author    = {Kim, Moo Jin and Pertsch, Karl and Karamcheti, Siddharth and Xiao, Ted and Balakrishna, Ashwin and Nair, Suraj and Rafailov, Rafael and Foster, Ethan and Sanketi, Pannag and Vuong, Quan and Kollar, Thomas and Burchfiel, Benjamin and Tedrake, Russ and Sadigh, Dorsa and Levine, Sergey and Liang, Percy and Finn, Chelsea},
	booktitle = {Conference on Robot Learning (CoRL)},
	series    = {Proceedings of Machine Learning Research},
	volume    = {270},
	pages     = {2679--2713},
	publisher = {PMLR},
	year      = {2024}
}

@inproceedings{pi0_2024,
	title     = {$\pi_0$: A Vision-Language-Action Flow Model for General Robot Control},
	author    = {Black, Kevin and others},
	booktitle = {Robotics: Science and Systems (RSS)},
	year      = {2025},
	doi       = {10.15607/RSS.2025.XXI.010}
}

@article{pi05_2025,
	title   = {$\pi_{0.5}$: A Vision-Language-Action Model with Open-World Generalization},
	author  = {{Physical Intelligence} and others},
	journal = {arXiv preprint arXiv:2504.16054},
	year    = {2025}
}

@article{gr00t2025,
	title   = {{GR00T N1}: An Open Foundation Model for Generalist Humanoid Robots},
	author  = {{NVIDIA} and others},
	journal = {arXiv preprint arXiv:2503.14734},
	year    = {2025},
	note    = {The {N1.5} checkpoint used by {U0} and in this paper is a later release of the same model line}
}

@article{smolvla2025,
	title   = {{SmolVLA}: A Vision-Language-Action Model for Affordable and Efficient Robotics},
	author  = {Shukor, Mustafa and Aubakirova, Dana and Capuano, Francesco and Kooijmans, Pepijn and Palma, Steven and Zouitine, Adil and Aractingi, Michel and Pascal, Caroline and Russi, Martino and Marafioti, Andres and Alibert, Simon and others},
	journal = {arXiv preprint arXiv:2506.01844},
	year    = {2025}
}

@inproceedings{xvla2026,
	title     = {{X-VLA}: Soft-Prompted Transformer as Scalable Cross-Embodiment Vision-Language-Action Model},
	author    = {Zheng, Jinliang and Li, Jianxiong and Wang, Zhihao and Liu, Dongxiu and Kang, Xirui and Feng, Yuchun and Zheng, Yinan and Zou, Jiayin and Chen, Yilun and Zeng, Jia and Zhang, Ya-Qin and Pang, Jiangmiao and Liu, Jingjing and Wang, Tai and Zhan, Xianyuan},
	booktitle = {International Conference on Learning Representations (ICLR)},
	year      = {2026},
	note      = {arXiv:2510.10274}
}

@article{ha2018worldmodels,
	title   = {World Models},
	author  = {Ha, David and Schmidhuber, J{\"u}rgen},
	journal = {arXiv preprint arXiv:1803.10122},
	year    = {2018}
}

@inproceedings{hafner2019planet,
	title     = {Learning Latent Dynamics for Planning from Pixels},
	author    = {Hafner, Danijar and Lillicrap, Timothy and Fischer, Ian and Villegas, Ruben and Ha, David and Lee, Honglak and Davidson, James},
	booktitle = {International Conference on Machine Learning (ICML)},
	series    = {Proceedings of Machine Learning Research},
	volume    = {97},
	pages     = {2555--2565},
	publisher = {PMLR},
	year      = {2019}
}

@inproceedings{hafner2020dreamer,
	title     = {Dream to Control: Learning Behaviors by Latent Imagination},
	author    = {Hafner, Danijar and Lillicrap, Timothy and Ba, Jimmy and Norouzi, Mohammad},
	booktitle = {International Conference on Learning Representations (ICLR)},
	year      = {2020},
	note      = {arXiv:1912.01603}
}

@article{hafner2023dreamerv3,
	title   = {Mastering Diverse Control Tasks through World Models},
	author  = {Hafner, Danijar and Pasukonis, Jurgis and Ba, Jimmy and Lillicrap, Timothy},
	journal = {Nature},
	volume  = {640},
	number  = {8059},
	pages   = {647--653},
	year    = {2025},
	doi     = {10.1038/s41586-025-08744-2}
}

@inproceedings{chua2018pets,
	title     = {Deep Reinforcement Learning in a Handful of Trials using Probabilistic Dynamics Models},
	author    = {Chua, Kurtland and Calandra, Roberto and McAllister, Rowan and Levine, Sergey},
	booktitle = {Advances in Neural Information Processing Systems (NeurIPS)},
	volume    = {31},
	year      = {2018}
}

@inproceedings{hansen2024tdmpc2,
	title     = {{TD-MPC2}: Scalable, Robust World Models for Continuous Control},
	author    = {Hansen, Nicklas and Su, Hao and Wang, Xiaolong},
	booktitle = {International Conference on Learning Representations (ICLR)},
	year      = {2024},
	note      = {arXiv:2310.16828}
}

@article{hu2023gaia1,
	title   = {{GAIA-1}: A Generative World Model for Autonomous Driving},
	author  = {Hu, Anthony and Russell, Lloyd and Yeo, Hudson and Murez, Zak and Fedoseev, George and Kendall, Alex and Shotton, Jamie and Corrado, Gianluca},
	journal = {arXiv preprint arXiv:2309.17080},
	year    = {2023}
}

@inproceedings{yang2023unisim,
	title     = {Learning Interactive Real-World Simulators},
	author    = {Yang, Sherry and Du, Yilun and Ghasemipour, Kamyar and Tompson, Jonathan and Kaelbling, Leslie and Schuurmans, Dale and Abbeel, Pieter},
	booktitle = {International Conference on Learning Representations (ICLR)},
	year      = {2024},
	note      = {arXiv:2310.06114}
}

@inproceedings{bruce2024genie,
	title     = {Genie: Generative Interactive Environments},
	author    = {Bruce, Jake and others},
	booktitle = {International Conference on Machine Learning (ICML)},
	series    = {Proceedings of Machine Learning Research},
	volume    = {235},
	pages     = {4603--4623},
	publisher = {PMLR},
	year      = {2024}
}

@article{nvidia2025cosmos,
	title   = {Cosmos World Foundation Model Platform for Physical {AI}},
	author  = {{NVIDIA} and others},
	journal = {arXiv preprint arXiv:2501.03575},
	year    = {2025}
}

@inproceedings{hu2024vpp,
	title     = {Video Prediction Policy: A Generalist Robot Policy with Predictive Visual Representations},
	author    = {Hu, Yucheng and Guo, Yanjiang and Wang, Pengchao and Chen, Xiaoyu and Wang, Yen-Jen and Zhang, Jianke and Sreenath, Koushil and Lu, Chaochao and Chen, Jianyu},
	booktitle = {International Conference on Machine Learning (ICML)},
	series    = {Proceedings of Machine Learning Research},
	volume    = {267},
	pages     = {24328--24346},
	publisher = {PMLR},
	year      = {2025}
}

@inproceedings{li2025uva,
	title     = {Unified Video Action Model},
	author    = {Li, Shuang and Gao, Yihuai and Sadigh, Dorsa and Song, Shuran},
	booktitle = {Robotics: Science and Systems (RSS)},
	year      = {2025},
	doi       = {10.15607/RSS.2025.XXI.074}
}

@article{dreamzero2026,
	title   = {World Action Models are Zero-shot Policies},
	author  = {Ye, Seonghyeon and others},
	journal = {arXiv preprint arXiv:2602.15922},
	year    = {2026}
}

@book{fossen2011handbook,
	title     = {Handbook of Marine Craft Hydrodynamics and Motion Control},
	author    = {Fossen, Thor I.},
	publisher = {John Wiley \& Sons},
	year      = {2011},
	doi       = {10.1002/9781119994138}
}

@article{yoerger1990thruster,
	title   = {The Influence of Thruster Dynamics on Underwater Vehicle Behavior and Their Incorporation into Control System Design},
	author  = {Yoerger, Dana R. and Cooke, John G. and Slotine, Jean-Jacques E.},
	journal = {IEEE Journal of Oceanic Engineering},
	volume  = {15},
	number  = {3},
	pages   = {167--178},
	year    = {1990},
	doi     = {10.1109/48.107145}
}

@article{bachmayer2000thruster,
	title   = {An Accurate Four-Quadrant Nonlinear Dynamical Model for Marine Thrusters: Theory and Experimental Validation},
	author  = {Bachmayer, Ralf and Whitcomb, Louis L. and Grosenbaugh, Mark A.},
	journal = {IEEE Journal of Oceanic Engineering},
	volume  = {25},
	number  = {1},
	pages   = {146--159},
	year    = {2000},
	doi     = {10.1109/48.820747}
}

@article{shen2018lmpc,
	title   = {Trajectory Tracking Control of an Autonomous Underwater Vehicle Using {Lyapunov}-Based Model Predictive Control},
	author  = {Shen, Chao and Shi, Yang and Buckham, Bradley},
	journal = {IEEE Transactions on Industrial Electronics},
	volume  = {65},
	number  = {7},
	pages   = {5796--5805},
	year    = {2018},
	doi     = {10.1109/TIE.2017.2779442}
}

@article{heshmati2020nmpc,
	title   = {A Robust Predictive Control Approach for Underwater Robotic Vehicles},
	author  = {Heshmati-Alamdari, Shahab and Karras, George C. and Marantos, Panos and Kyriakopoulos, Kostas J.},
	journal = {IEEE Transactions on Control Systems Technology},
	volume  = {28},
	number  = {6},
	pages   = {2352--2363},
	year    = {2020},
	doi     = {10.1109/TCST.2019.2939248}
}

@inproceedings{wehbe2017identification,
	title     = {Experimental Evaluation of Various Machine Learning Regression Methods for Model Identification of Autonomous Underwater Vehicles},
	author    = {Wehbe, Bilal and Hildebrandt, Marc and Kirchner, Frank},
	booktitle = {IEEE International Conference on Robotics and Automation (ICRA)},
	pages     = {4885--4890},
	year      = {2017},
	doi       = {10.1109/ICRA.2017.7989565}
}

@article{carlucho2018drl,
	title   = {Adaptive Low-Level Control of Autonomous Underwater Vehicles Using Deep Reinforcement Learning},
	author  = {Carlucho, Ignacio and De Paula, Mariano and Wang, Sen and Petillot, Yvan and Acosta, Gerardo G.},
	journal = {Robotics and Autonomous Systems},
	volume  = {107},
	pages   = {71--86},
	year    = {2018},
	doi     = {10.1016/j.robot.2018.05.016}
}

@inproceedings{kinsey2006survey,
	title     = {A Survey of Underwater Vehicle Navigation: Recent Advances and New Challenges},
	author    = {Kinsey, James C. and Eustice, Ryan M. and Whitcomb, Louis L.},
	booktitle = {IFAC Conference on Manoeuvring and Control of Marine Craft (MCMC)},
	year      = {2006}
}

@article{sivcev2018manipulators,
	title   = {Underwater Manipulators: A Review},
	author  = {Siv{\v{c}}ev, Satja and Coleman, Joseph and Omerdi{\'c}, Edin and Dooly, Gerard and Toal, Daniel},
	journal = {Ocean Engineering},
	volume  = {163},
	pages   = {431--450},
	year    = {2018},
	doi     = {10.1016/j.oceaneng.2018.06.018}
}

@article{paull2014review,
	title   = {{AUV} Navigation and Localization: A Review},
	author  = {Paull, Liam and Saeedi, Sajad and Seto, Mae and Li, Howard},
	journal = {IEEE Journal of Oceanic Engineering},
	volume  = {39},
	number  = {1},
	pages   = {131--149},
	year    = {2014},
	doi     = {10.1109/JOE.2013.2278891}
}

@inproceedings{topini2020lstm,
	title     = {{LSTM}-based Dead Reckoning Navigation for Autonomous Underwater Vehicles},
	author    = {Topini, Edoardo and Topini, Alberto and Franchi, Matteo and Bucci, Alessandro and Secciani, Nicola and Ridolfi, Alessandro and Allotta, Benedetto},
	booktitle = {Global Oceans 2020: Singapore -- U.S. Gulf Coast},
	pages     = {1--7},
	publisher = {IEEE},
	year      = {2020},
	doi       = {10.1109/IEEECONF38699.2020.9389379}
}

@inproceedings{cohen2023setbeamsnet,
	title     = {{Set-Transformer BeamsNet} for {AUV} Velocity Forecasting in Complete {DVL} Outage Scenarios},
	author    = {Cohen, Nadav and Yampolsky, Zeev and Klein, Itzik},
	booktitle = {IEEE Underwater Technology (UT)},
	pages     = {1--6},
	year      = {2023},
	doi       = {10.1109/UT49729.2023.10103453}
}

@inproceedings{cieslak2019stonefish,
	title     = {Stonefish: An Advanced Open-Source Simulation Tool Designed for Marine Robotics, With a {ROS} Interface},
	author    = {Cie{\'s}lak, Patryk},
	booktitle = {OCEANS 2019 -- Marseille},
	pages     = {1--6},
	year      = {2019},
	doi       = {10.1109/OCEANSE.2019.8867434}
}

@inproceedings{ross2011dagger,
	title     = {A Reduction of Imitation Learning and Structured Prediction to No-Regret Online Learning},
	author    = {Ross, St{\'e}phane and Gordon, Geoffrey J. and Bagnell, J. Andrew},
	booktitle = {Proceedings of the Fourteenth International Conference on Artificial Intelligence and Statistics (AISTATS)},
	series    = {Proceedings of Machine Learning Research},
	volume    = {15},
	pages     = {627--635},
	publisher = {PMLR},
	year      = {2011}
}

@article{oquab2024dinov2,
	title   = {{DINOv2}: Learning Robust Visual Features without Supervision},
	author  = {Oquab, Maxime and Darcet, Timoth{\'e}e and Moutakanni, Th{\'e}o and Vo, Huy V. and Szafraniec, Marc and Khalidov, Vasil and Fernandez, Pierre and Haziza, Daniel and Massa, Francisco and El-Nouby, Alaaeldin and Assran, Mido and Ballas, Nicolas and Galuba, Wojciech and Howes, Russell and Huang, Po-Yao and Li, Shang-Wen and Misra, Ishan and Rabbat, Michael and Sharma, Vasu and Synnaeve, Gabriel and Xu, Hu and J{\'e}gou, Herv{\'e} and Mairal, Julien and Labatut, Patrick and Joulin, Armand and Bojanowski, Piotr},
	journal = {Transactions on Machine Learning Research},
	year    = {2024}
}

@inproceedings{lynch2019play,
	title     = {Learning Latent Plans from Play},
	author    = {Lynch, Corey and Khansari, Mohi and Xiao, Ted and Kumar, Vikash and Tompson, Jonathan and Levine, Sergey and Sermanet, Pierre},
	booktitle = {Conference on Robot Learning (CoRL)},
	series    = {Proceedings of Machine Learning Research},
	volume    = {100},
	pages     = {1113--1132},
	publisher = {PMLR},
	year      = {2019}
}

@inproceedings{cho2014gru,
	title     = {Learning Phrase Representations using {RNN} Encoder--Decoder for Statistical Machine Translation},
	author    = {Cho, Kyunghyun and van Merri{\"e}nboer, Bart and Gulcehre, Caglar and Bahdanau, Dzmitry and Bougares, Fethi and Schwenk, Holger and Bengio, Yoshua},
	booktitle = {Conference on Empirical Methods in Natural Language Processing (EMNLP)},
	pages     = {1724--1734},
	year      = {2014},
	doi       = {10.3115/v1/D14-1179}
}

@article{anderson2018spl,
	title   = {On Evaluation of Embodied Navigation Agents},
	author  = {Anderson, Peter and Chang, Angel and Chaplot, Devendra Singh and Dosovitskiy, Alexey and Gupta, Saurabh and Koltun, Vladlen and Kosecka, Jana and Malik, Jitendra and Mottaghi, Roozbeh and Savva, Manolis and Zamir, Amir R.},
	journal = {arXiv preprint arXiv:1807.06757},
	year    = {2018}
}

@article{uhlenbeck1930ou,
	title   = {On the Theory of the {B}rownian Motion},
	author  = {Uhlenbeck, George E. and Ornstein, Leonard S.},
	journal = {Physical Review},
	volume  = {36},
	number  = {5},
	pages   = {823--841},
	year    = {1930},
	doi     = {10.1103/PhysRev.36.823}
}

@book{jerlov1976marine,
	title     = {Marine Optics},
	author    = {Jerlov, Nils G.},
	series    = {Elsevier Oceanography Series},
	volume    = {14},
	edition   = {2nd},
	publisher = {Elsevier},
	address   = {Amsterdam},
	year      = {1976}
}

@article{hendrycks2016gelu,
	title   = {Gaussian Error Linear Units ({GELUs})},
	author  = {Hendrycks, Dan and Gimpel, Kevin},
	journal = {arXiv preprint arXiv:1606.08415},
	year    = {2016}
}

@article{ba2016layernorm,
	title   = {Layer Normalization},
	author  = {Ba, Jimmy Lei and Kiros, Jamie Ryan and Hinton, Geoffrey E.},
	journal = {arXiv preprint arXiv:1607.06450},
	year    = {2016}
}

@article{srivastava2014dropout,
	title   = {Dropout: A Simple Way to Prevent Neural Networks from Overfitting},
	author  = {Srivastava, Nitish and Hinton, Geoffrey and Krizhevsky, Alex and Sutskever, Ilya and Salakhutdinov, Ruslan},
	journal = {Journal of Machine Learning Research},
	volume  = {15},
	number  = {56},
	pages   = {1929--1958},
	year    = {2014}
}

@article{cen2025worldvla,
	title = {{WorldVLA}: Towards Autoregressive Action World Model},
	author = {Cen, Jun and Yu, Chaohui and Yuan, Hangjie and Jiang, Yuming and Huang, Siteng and Guo, Jiayan and Li, Xin and Song, Yibing and Luo, Hao and Wang, Fan and Zhao, Deli and Chen, Hao},
	journal = {arXiv preprint arXiv:2506.21539},
	year = {2025}
}

@inproceedings{kim2026cosmospolicy,
	title     = {{Cosmos Policy}: Fine-Tuning Video Models for Visuomotor Control and Planning},
	author    = {Kim, Moo Jin and Gao, Yihuai and Lin, Tsung-Yi and Lin, Yen-Chen and Ge, Yunhao and Lam, Grace and Liang, Percy and Song, Shuran and Liu, Ming-Yu and Finn, Chelsea and Gu, Jinwei},
	booktitle = {International Conference on Learning Representations (ICLR)},
	year      = {2026},
	note      = {arXiv:2601.16163}
}

@inproceedings{li2026lingbotva,
	title     = {Causal World Modeling for Robot Control},
	author    = {Li, Lin and Zhang, Qihang and Luo, Yiming and Yang, Shuai and Wang, Ruilin and Zhang, Luyao and Yu, Mingrui and Gao, Zelin and Xue, Nan and Zhou, Boyu and Zhu, Xing and Ding, Mingyu and Shen, Yujun and Xu, Yinghao},
	booktitle = {Robotics: Science and Systems (RSS)},
	year      = {2026},
	doi       = {10.15607/RSS.2026.XXII.016}
}

@techreport{karumbunathan2022orin,
	title       = {{NVIDIA Jetson AGX Orin Series}: A Giant Leap Forward for Robotics and Edge {AI} Applications},
	author      = {Karumbunathan, Leela S.},
	institution = {NVIDIA},
	year        = {2022},
	month       = jul,
	type        = {Technical Brief},
	number      = {TB\_10749-001\_v1.2},
	url         = {https://www.nvidia.com/content/dam/en-zz/Solutions/gtcf21/jetson-orin/nvidia-jetson-agx-orin-technical-brief.pdf}
}
	
	\newpage
	\appendix
	\raggedbottom
	\let\oldsection\section
	\renewcommand{\section}{\Needspace{6\baselineskip}\oldsection}
	\setlength{\intextsep}{8pt plus 2pt minus 2pt}
	\setlength{\abovecaptionskip}{3pt}
	\section*{Appendix}
	
	\section{The Predictor}
	\label{app:implementation}
	
	\begin{wraptable}{r}{0.50\textwidth}
		\vspace{-1.35\baselineskip}
		\centering
		\setlength{\abovecaptionskip}{1pt}
		\setlength{\belowcaptionskip}{1pt}
		\footnotesize
		\setlength{\tabcolsep}{3pt}
		\renewcommand{\arraystretch}{1.02}
		\caption{\textbf{Details of the 35-dimensional physical state used in AquaWAM.}}
		\label{tab:state}
		\begin{tabular}{@{}l c l@{}}
			\toprule
			Channels & Dim. & Sensor \\
			\midrule
			\multicolumn{3}{@{}l}{\textit{Vehicle (19)}} \\
			body velocity $v$ & 3 & DVL \\
			angular rate $\omega$ & 3 & IMU \\
			linear acceleration & 3 & IMU \\
			depth, altitude & 2 & pressure, DVL \\
			last thruster command & 8 & own output \\
			\midrule
			\multicolumn{3}{@{}l}{\textit{Arm and object (16)}} \\
			joint positions and velocities & 10 & encoders \\
			object pose in the gripper frame & 6 & wrist camera \\
			\bottomrule
		\end{tabular}
	\end{wraptable}
	Table~\ref{tab:state} lists the channels of the physical state, all read from the vehicle's own sensors. Specifically, the 19-dimensional vehicle block consists of body velocity and altitude from the DVL, angular rate and linear acceleration from the IMU, depth from the pressure sensor, and the last thruster command. Within reach of an object, the joint encoders and the wrist-camera head add 16 further dimensions, and the predictor predicts every channel. Each command consists of eight thruster values in $[-1, 1]$ and five joint targets, the first being the gripper (0 open, 0.015 closed). The history encoder $E_\phi$ embeds the last 16 steps (1.6~s) of states and commands with a linear layer and a Gaussian error linear unit (GELU) \citep{hendrycks2016gelu}, runs a single-layer gated recurrent unit (GRU) \citep{cho2014gru} with a hidden width of 384, and maps its last hidden state through layer normalization \citep{ba2016layernorm}, dropout \citep{srivastava2014dropout} and a linear layer to the 96-dimensional disturbance token; its parameters $\phi$ are part of $\theta$ and are trained jointly with $f_{\mathrm{nom}}$ and $f_{\mathrm{res}}$. The nominal and residual networks are three-layer multilayer perceptrons (MLPs) of the same width, with GELU activations, layer normalization and a dropout of 0.1. Each call predicts $K = 5$ steps, and four chained calls cover the two-second horizon. The velocity estimate is the mean of three heads that read the history with the velocity channels masked, and their spread gives $\sigma_t$. Both scales of the state share this network, differing only in input and output widths. The perception heads, built on DINOv2-base, never enter the predictor, and the gripper closes when an outcome head on the predicted state reports a probability of at least 0.7.
	
	\begin{algorithm}[h]
		\caption{One decision of \method{} (every half second)}
		\label{alg:decision}
		\small
		\begin{algorithmic}[1]
			\Require state $s_t$ from the DVL, IMU, pressure sensor and joint encoders; history $\mathcal{H}_t$; camera images; instruction
			\Ensure the future $\hat{s}^{\star}_{t+1:t+H}$ that \method{} has chosen and the commands $a^{\star}_{t:t+K}$ that bring it about (Eq.~\ref{eq:wam})
			\State goal $g$, target pose $\gets$ perception heads(camera images, instruction) \Comment{Section~\ref{sec:method-perception}}
			\If{the DVL has lost bottom lock} \Comment{Section~\ref{sec:method-estimator}}
			\State $\hat{v}_t \gets \estimate(\mathcal{H}_t^{v})$; write $\hat{v}_t$ into $s_t$; dead-reckon the pose; update the trust $\alpha_t$ (Eq.~\ref{eq:trust})
			\EndIf
			\State scale $\gets$ vehicle block while cruising, full state within 10~cm of the object \Comment{Section~\ref{sec:method-state}}
			\State $\mathcal{A}_t \gets$ thrust laws, single-axis moves and play snippets while cruising; thrust pulses within reach (Eq.~\ref{eq:pulses})
			\For{each candidate $\mathbf{a} \in \mathcal{A}_t$, in one batch}
			\State $\hat{s}(\mathbf{a}) \gets \imagine(s_t, \mathcal{H}_t, \mathbf{a})$ \Comment{Eq.~\ref{eq:predictor} chained four times, two seconds ahead}
			\EndFor
			\State while cruising: draw a second round around the 16 best and imagine it too
			\State $\mathbf{a}^{\star} \gets \arg\min_{\mathbf{a}} J(\hat{s}(\mathbf{a}))$ \Comment{Eq.~\ref{eq:score}; $J^{\mathrm{pulse}}$ within reach}
			\State execute the first half second of $\mathbf{a}^{\star}$; close the gripper if $p(\text{grasp} \mid \hat{s}(\mathbf{a}^{\star})) \ge 0.7$
		\end{algorithmic}
	\end{algorithm}
	
	\section{Scores and the Hand-Tuned Law}
	\label{app:scores}
	
	The weights of the score in Eq.~\ref{eq:score} are $\lambda_g = 5$, $\lambda_u = 0.02$, $\lambda_s = 0.4$ and $\lambda_\omega = 0.5$. The speed cap $v_{\max}$ follows the current task: 0.25, 0.22 and 0.20~m/s for goto, scan and inspection, and 0.25~m/s when following the boat. Because the cap enters only as a soft penalty, the goal term outweighs it during tracking, which allows the vehicle to match the boat's 0.44~m/s. The velocity weight of the pulse score in Eq.~\ref{eq:pulses} is $\lambda_v = 400$, and the noise floor of the trust statistic in Eq.~\ref{eq:trust} is $\sigma_0 = 0.05$, in units of the normalized velocity.
	
	The pulse candidates of Eq.~\ref{eq:pulses} combine the magnitudes $m \in \{0.12, 0.18, 0.25, 0.32, 0.40\}$, in units of the normalized body-thrust command that the mixer maps to the eight thruster values in $[-1, 1]$, with the durations $\tau \in \{0.1, 0.2, 0.3, 0.4\}$~s, i.e., at most $K-1$ steps, because the last executed step is always a coast. This $5\times4$ grid (Section~\ref{sec:results-manipulation}) yields 20 pulses per direction; the coarse grid there is $\{0.18, 0.32\} \times \{0.2, 0.4\}$~s and the fine grid has nine magnitudes. An axis direction $\operatorname{sign}(\delta_i)\,\mathbf{b}_i$ enters only when $|\delta_i| > 4$~mm, and the braking direction $-v_t/\lVert v_t \rVert$, with $v_t$ the body velocity measured by the DVL, only when $\lVert v_t \rVert > 1.5$~cm/s, so a decision weighs one to five directions, that is, between 21 and 101 candidates. There is no separate braking pulse: braking is a pulse along $-v_t$. The hold candidate issues no translational thrust over the whole horizon and keeps only the thrust that levels the attitude. In $J^{\mathrm{pulse}}$, the offset $e$ is taken in the world frame, with $x$ and $y$ horizontal as in the judge and $z$ vertical, measured from a target point 1.2~cm above the object; $\lVert \hat{e}_{\mathrm{end}} \rVert^2_Q$ is the mean of $e^{\top} Q e$ over the last $\lfloor H/3 \rfloor = 6$ predicted steps (1.5 to 2.0~s), and $\hat{v}_{\mathrm{end}}$ is the predicted linear velocity at step $t+H$.
	
	The hand-tuned law of Section~\ref{sec:results-manipulation} shares the task program and the closure window of \method{}, so that the two differ only in how the hull is positioned. It commands a body speed of 0.03~m/s, held for $\mathrm{distance}/(0.03 \times 0.1~\mathrm{s})$ steps, and switches from approach to grasp at 8~cm. It closes the gripper once $|\Delta y| < 0.7$~cm and $|\Delta x| < 3$~cm, where $\Delta x$ and $\Delta y$ are the horizontal offsets from the gripper to the object rotated into the heading frame of the vehicle ($x$ forward, $y$ sideways), and once the distance is below 3.5~cm, the yaw error is below 0.2~rad and the hull is at rest.
	
	\section{Tasks}
	\label{app:schema}
	
	\begin{wraptable}[15]{r}{0.54\textwidth}
		\vspace{-1.15\baselineskip}
		\centering
		\setlength{\abovecaptionskip}{1pt}
		\setlength{\belowcaptionskip}{2pt}
		\footnotesize
		\setlength{\tabcolsep}{3.5pt}
		\renewcommand{\arraystretch}{0.93}
		\caption{\textbf{The twenty USIM tasks} and trials per task.}
		\label{tab:tasks}
		\begin{tabular}{@{}l c @{\hspace{0.7em}} l c@{}}
			\toprule
			Task & n & Task & n \\
			\midrule
			\multicolumn{4}{@{}l}{\textit{Navigation}} \\
			goto charge station & 40 & scan ship modern & 20 \\
			goto water tower & 40 & scan ship ancient & 20 \\
			inspect pipeline pool & 20 & inspect pipeline sea & 20 \\
			\midrule
			\multicolumn{4}{@{}l}{\textit{Grasping}} \\
			pick pipe0 shallow & 40 & pick pipe1 shallow & 40 \\
			pick pipe0 factory & 40 & pick pipe1 factory & 40 \\
			pick red shallow & 40 & pick redx shallow & 40 \\
			pick red factory & 40 & pick redx factory & 40 \\
			pick blue shallow & 40 & pick bluex shallow & 40 \\
			pick blue factory & 40 & pick bluex factory & 40 \\
			\midrule
			\multicolumn{2}{@{}l}{\textit{Transporting}} & \multicolumn{2}{@{}l}{\textit{Tracking}} \\
			transfer red shallow & 40 & follow boat & 20 \\
			\bottomrule
		\end{tabular}
	\end{wraptable}
	Table~\ref{tab:tasks} lists the twenty USIM tasks and the number of trials behind each entry of Table~\ref{tab:main}. \method{} follows USIM's task schema, which orders the sub-goals, and advances to the next waypoint once the hull is within 0.9~m of it for a goto, within 3.5~m and 0.8~rad for a scan node, and within 1.5~m and 0.4~rad for an inspection node; all of these thresholds lie inside the judge's own tolerances. The goal velocity points at the current waypoint and is capped as in Appendix~\ref{app:scores}. When following the boat, the standoff is 3~m behind it along its heading and 0.5~m deeper. Within reach of an object, the arm holds a ready pose with joint targets $(0, 0.50, 0.50, 0.51, 0)$, the first of which keeps the gripper open, while the hull performs the positioning, and the pulse candidates take over within 10~cm. On the decision of the outcome head, the gripper is commanded to its closed position (joint target 0.015), the same value the experts use in the demonstrations, after which a transport task carries the object to the box head's goal.
	
	\section{Play Data}
	\label{app:play}
	
	The demonstrations consist of repeated task executions, whereas the play carries no task or success label. The play consists of three recordings of the same vehicle, namely an Ornstein--Uhlenbeck exploration \citep{uhlenbeck1930ou}, a pass through the scenes and a deployment log. Because its commands differ from those the experts issue, the play exposes the predictor to the dead band and to the glide beyond it, which the demonstrations alone do not reveal (Section~\ref{sec:results-ablations}).
	
	\section{Baselines and DVL Dropout}
	\label{app:baselines}
	
	\paragraph{Fine-tuning.} Every model in Table~\ref{tab:main} is fine-tuned on the same demonstrations. U0 trains its action head and commits to a 16-step chunk. OpenVLA uses sandwich fine-tuning, in which the vision encoders, the projector and the output layer are trained while the language model stays frozen, for 22k steps at batch size 64, with the actions discretized into 256 bins. GR00T~N1.5 receives full action-head fine-tuning for 120k steps at batch size 16 and, like U0, commits to a 16-step chunk. $\pi_{0.5}$ trains the vision tower and the action expert with the language model frozen, on absolute actions, for 22k steps at batch size 64, and, like OpenVLA, acts at every step. X-VLA adapts its released base model for 120k steps at batch size 16 and renews its chunk every 16 steps, as does SmolVLA, which is fine-tuned for 65k steps at batch size 32.
	
	\paragraph{DVL dropout.} Table~\ref{tab:dropout-tasks} compares the per-task success of \method{} under full sensing with that under the dropout of Section~\ref{sec:results-dropout}. The goto, scan and inspection tasks are almost unaffected, and tracking drops from 20 to 16 of 20 trials. In contrast, grasping loses at most 12 of 40 successes per object, and transport falls from 23 to 3 of 40, since the object must still be carried after the loss.
	
	\begin{table}[t]
		\caption{\textbf{Per-task success of \method{} with full sensing and under DVL dropout.} Bold marks the higher of the two.}
		\label{tab:dropout-tasks}
		\centering
		\small
		\setlength{\tabcolsep}{5pt}
		\renewcommand{\arraystretch}{1.06}
		\begin{tabular*}{\textwidth}{@{}l @{\extracolsep{\fill}} cc @{\extracolsep{1.4em}} l cc@{}}
			\toprule
			Task & Full & Dropout & Task & Full & Dropout \\
			\midrule
			goto charge station & 40/40 & 40/40 & scan ship modern & 19/20 & \textbf{20/20} \\
			goto water tower & \textbf{36/40} & 33/40 & scan ship ancient & 16/20 & \textbf{17/20} \\
			inspect pipeline pool & 20/20 & 20/20 & inspect pipeline sea & 20/20 & 20/20 \\
			follow boat & \textbf{20/20} & 16/20 & transfer red shallow & \textbf{23/40} & 3/40 \\
			\midrule
			pick pipe0 shallow & \textbf{33/40} & 21/40 & pick pipe1 shallow & \textbf{33/40} & 25/40 \\
			pick pipe0 factory & \textbf{28/40} & 21/40 & pick pipe1 factory & \textbf{26/40} & 25/40 \\
			pick red shallow & \textbf{24/40} & 21/40 & pick redx shallow & 24/40 & \textbf{25/40} \\
			pick red factory & \textbf{22/40} & 17/40 & pick redx factory & 20/40 & \textbf{21/40} \\
			pick blue shallow & \textbf{34/40} & 27/40 & pick bluex shallow & \textbf{22/40} & 18/40 \\
			pick blue factory & \textbf{28/40} & 21/40 & pick bluex factory & 20/40 & 20/40 \\
			\bottomrule
		\end{tabular*}
	\end{table}
	
	\begin{table}[t]
		\caption{\textbf{The estimator of \method{} as a takeover layer for U0 when DVL is not available.} Bold marks the best in each column.}
		\label{tab:takeover}
		\centering
		\small
		\renewcommand{\arraystretch}{1.06}
		\begin{tabular*}{\textwidth}{@{}l @{\extracolsep{\fill}} ccccc@{}}
			\toprule
			Model & Navigation & Grasping & Transporting & Tracking & Overall SR \\
			\midrule
			U0 & 73/160 & 202/480 & 0/40 & 1/20 & 39.4\% (276/700) \\
			U0 + takeover & \textbf{155/160} & 181/480 & 0/40 & 12/20 & 49.7\% (348/700) \\
			\method{} & 150/160 & \textbf{262/480} & \textbf{3/40} & \textbf{16/20} & \textbf{61.6\% (431/700)} \\
			\bottomrule
		\end{tabular*}
	\end{table}
	
	\paragraph{Takeover layer.} Table~\ref{tab:takeover} breaks down by family the takeover experiment of Section~\ref{sec:results-dropout}, in which, once the DVL reports loss, U0 receives the velocity estimated by \method{} in place of the DVL measurement. Navigation rises from 73 to 155 of 160 and tracking from 1 to 12 of 20, whereas grasping falls from 202 to 181 of 480 and transport stays at 0 of 40.
	
	\section{USIM-Hard}
	\label{app:hard}
	
	Tables~\ref{tab:hard-tasks} and~\ref{tab:hard-strict} break the protocols of Table~\ref{tab:hard} down by task. Every perturbation is applied before the scene loads and is seeded by task and trial, so that every model faces the same perturbation in the same trial. The perturbations are defined as follows. \emph{Heading}: the start yaw is rotated by 90 to 180 degrees, whereas the demonstrations always start facing the goal. \emph{Tight endpoint}: the goto judge's ball shrinks from 1.0 to 0.5~m. \emph{Thruster at 50\%}: one horizontal thruster delivers half of its commanded thrust without announcement; the efficiency of thruster 3 falls inside the range covered by the play data, and that of thruster 1 outside it. \emph{Thruster 1 at 0\%}: that thruster delivers none of its commanded thrust. \emph{Turbid water}: the Stonefish water-type parameter, which follows the Jerlov classification \citep{jerlov1976marine} from clear oceanic water (0) to turbid coastal water (1), is raised from 0.15 to 0.50. The stricter judges, in contrast, perturb nothing: they re-score the recorded trajectories of Table~\ref{tab:main} and, in the two DVL-loss rows, those of the dropout runs of Section~\ref{sec:results-dropout}. For a goto, every intermediate waypoint must be passed within the stated radius; for an inspection, every node must lie within a 1~m ball and 90\% of the nodes must be covered.
	
	\begin{table}[t]
		\centering
		\small
		\setlength{\tabcolsep}{4pt}
		\renewcommand{\arraystretch}{1.06}
		\begin{minipage}[t]{0.485\textwidth}
			\caption{\textbf{Perturbed worlds of Table~\ref{tab:hard}, per task.} 20 trials per cell.}
			\label{tab:hard-tasks}
			\begin{tabular*}{\linewidth}{@{}l @{\extracolsep{\fill}} cc@{}}
				\toprule
				Task & U0 & \method{} \\
				\midrule
				\multicolumn{3}{@{}l}{\textit{Heading rotated 90--180$^\circ$}} \\
				goto charge station & 10/20 & 7/20 \\
				goto water tower & 16/20 & 16/20 \\
				inspect pipeline pool & 12/20 & 20/20 \\
				scan ship modern & 14/20 & 20/20 \\
				\midrule
				\multicolumn{3}{@{}l}{\textit{Goto endpoint within 0.5 m}} \\
				goto charge station & 18/20 & 20/20 \\
				goto water tower & 8/20 & 15/20 \\
				\midrule
				\multicolumn{3}{@{}l}{\textit{Thruster 3 at 50\%}} \\
				goto charge station & 3/20 & 18/20 \\
				goto water tower & 0/20 & 12/20 \\
				\midrule
				\multicolumn{3}{@{}l}{\textit{Thruster 1 at 50\%}} \\
				goto charge station & 4/20 & 13/20 \\
				goto water tower & 0/20 & 8/20 \\
				\midrule
				\multicolumn{3}{@{}l}{\textit{Thruster 1 at 0\%}} \\
				goto charge station & 1/20 & 4/20 \\
				goto water tower & 0/20 & 0/20 \\
				\midrule
				\multicolumn{3}{@{}l}{\textit{Turbid water, Jerlov 0.50}} \\
				goto water tower & 1/20 & 9/20 \\
				
				\bottomrule
			\end{tabular*}
		\end{minipage}\hfill
		\begin{minipage}[t]{0.485\textwidth}
			\caption{\textbf{Stricter judges of Table~\ref{tab:hard}, per task.} Gotos of 40 trials, inspections of 20.}
			\label{tab:hard-strict}
			\begin{tabular*}{\linewidth}{@{}l @{\extracolsep{\fill}} cc@{}}
				\toprule
				Task & U0 & \method{} \\
				\midrule
				\multicolumn{3}{@{}l}{\textit{Goto, all waypoints within 1.0 m}} \\
				goto charge station & 37/40 & 39/40 \\
				goto water tower & 16/40 & 22/40 \\
				\midrule
				\multicolumn{3}{@{}l}{\textit{Goto, all waypoints within 0.5 m}} \\
				goto charge station & 34/40 & 32/40 \\
				goto water tower & 10/40 & 12/40 \\
				\midrule
				\multicolumn{3}{@{}l}{\textit{Goto, 0.5 m waypoints, DVL loss}} \\
				goto charge station & 27/40 & 36/40 \\
				goto water tower & 1/40 & 9/40 \\
				\midrule
				\multicolumn{3}{@{}l}{\textit{Goto, all waypoints within 0.25 m}} \\
				goto charge station & 20/40 & 18/40 \\
				goto water tower & 5/40 & 6/40 \\
				\midrule
				\multicolumn{3}{@{}l}{\textit{Inspect, 1 m ball, 90\% coverage}} \\
				inspect pipeline pool & 16/20 & 19/20 \\
				inspect pipeline sea & 20/20 & 20/20 \\
				\midrule
				\multicolumn{3}{@{}l}{\textit{Inspect, same judge, DVL loss}} \\
				inspect pipeline pool & 4/20 & 20/20 \\
				inspect pipeline sea & 3/20 & 19/20 \\
				
				\bottomrule
			\end{tabular*}
		\end{minipage}
	\end{table}
	
	\section{Failure Cases}
	\label{app:failures}
	
	Three results in the main text mark where the margin of \method{} shrinks or where both models fail. We collect them here together with their causes.
	
	First, transport after DVL loss succeeds in only 3 of 40 trials, against 0 of 40 for U0. The object must still be carried 2.9~m after the loss, so the estimate, rather than a measurement, has to hold the velocity over the whole carry.
	
	Second, with thruster 1 delivering no thrust, a fault that the play never contains, the two goto tasks succeed in only 4 of 40 trials, against 1 of 40 for U0. A weakened thruster is largely absorbed by the disturbance token (21 of 40 at half thrust), whereas a missing one is not.
	
	Finally, turbid water with a Jerlov water-type parameter of 0.50 degrades the cameras that supply the goal, and \method{} reaches the water tower in only 9 of 20 trials, against 1 of 20 for U0.
	
	\clearpage
	\section{Qualitative Comparisons}
	\label{app:qualitative}
	
\begin{figure}[h]
		\centering
		\includegraphics[width=0.9\textwidth]{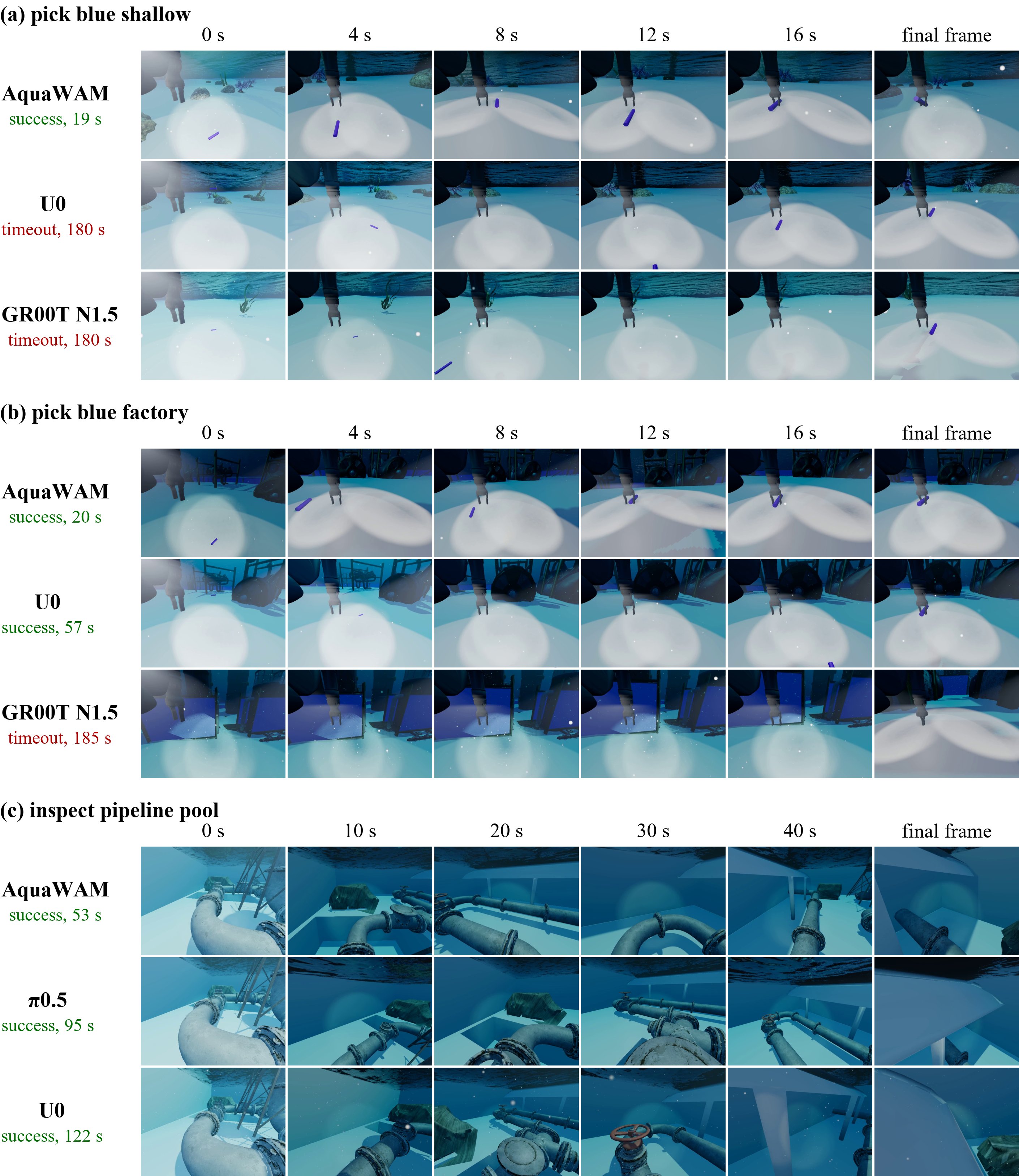}
		\caption{\textbf{Shared episodes: two grasps and an inspection.} Each panel follows one episode from identical initial conditions; the last column is the final frame.}
		\label{fig:qualitative}
	\end{figure}

\begin{figure}[p]
		\centering
		\includegraphics[width=0.9\textwidth]{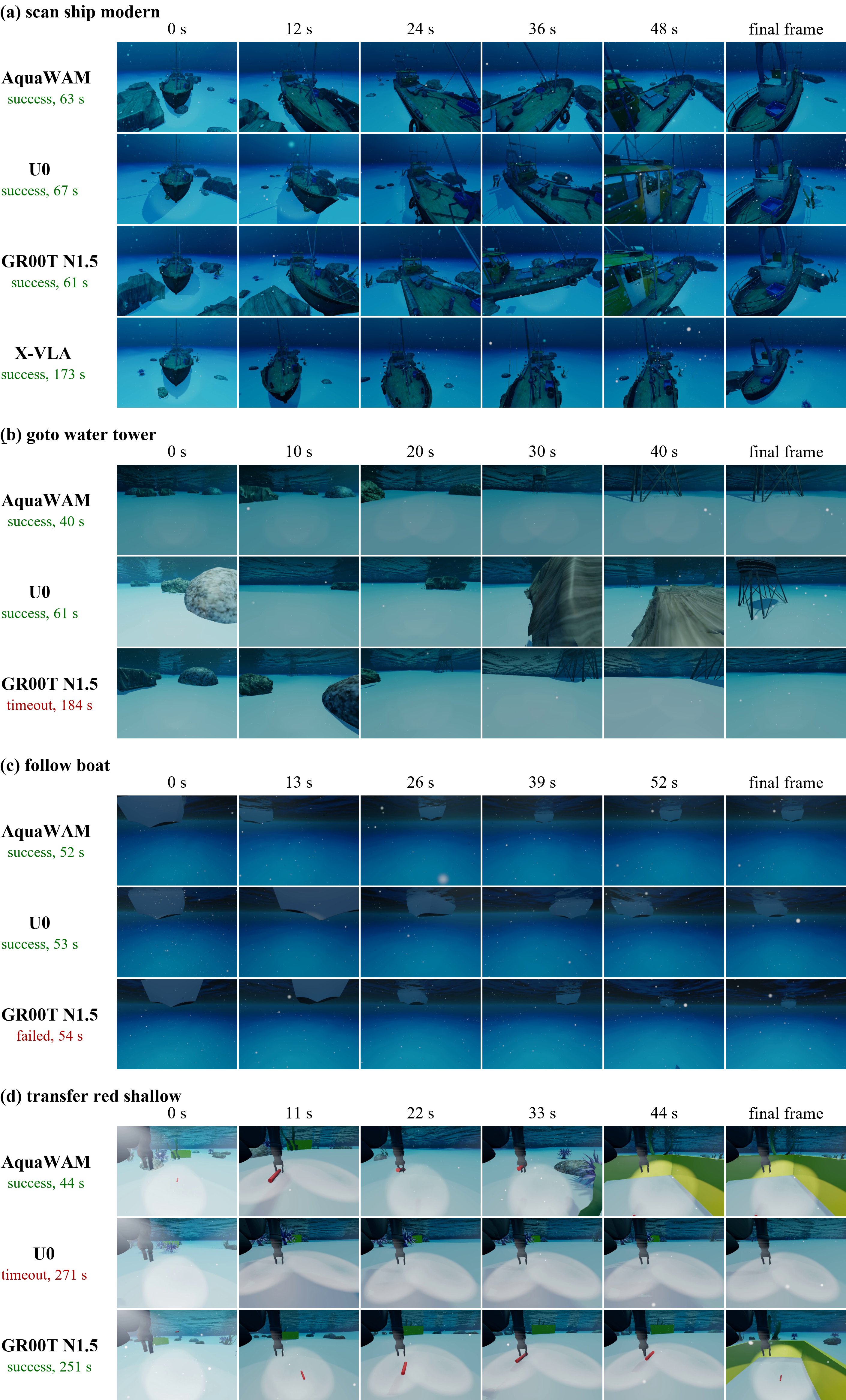}
		\caption{\textbf{Scanning, navigation, tracking and transport.} (a) Shared episode; in (b)--(d), columns mark quarters of the \method{} episode. The last column is the final frame.}
		\label{fig:qualitative2}
	\end{figure}

	Figures~\ref{fig:qualitative} and~\ref{fig:qualitative2} show selected episodes. In Figure~\ref{fig:qualitative} and Figure~\ref{fig:qualitative2}a, every model starts from identical initial conditions. On pick blue shallow, \method{} closes the gripper on the object at 19~s, whereas U0 and GR00T~N1.5 stay near it without a successful grasp until the 180~s timeout; on pick blue factory, \method{} succeeds at 20~s and U0 at 57~s, whereas GR00T~N1.5 times out. On inspect pipeline pool, \method{}, $\pi_{0.5}$ and U0 complete the route in 53, 95 and 122~s, and on scan ship modern, GR00T~N1.5, \method{} and U0 complete the scan in 61, 63 and 67~s and X-VLA in 173~s. Across these four episodes, $\pi_{0.5}$ and X-VLA succeed only on the inspection and the scan, respectively, and SmolVLA and OpenVLA on none. Figure~\ref{fig:qualitative2}b--d covers navigation to a landmark, tracking and transport, for which no shared episode was recorded; each row shows the episode of \method{} when it was also recorded for that model and otherwise the recorded episode of median duration, so these rows do not share an initial condition and their durations are not directly comparable. The supplementary video plays the shared episodes side by side, with the simulated time and the playback speed shown. All clips are selected examples, and the success rates are those of Tables~\ref{tab:main} and~\ref{tab:dropout-tasks}.
	
\end{document}